%% file: main.tex
\documentclass[sigconf]{acmart}

\AtBeginDocument{%
  }

\usepackage{multirow}
\usepackage{hyperref}
\usepackage[nameinlink,capitalise]{cleveref}

\usepackage{enumitem}
\setlist[enumerate]{noitemsep, topsep=0pt, parsep=0pt, partopsep=0pt}
\usepackage{csquotes}
\usepackage{graphicx}
\usepackage{float}
\usepackage{xcolor}
\usepackage{placeins}
\usepackage{subcaption}
\usepackage{algorithm}
\usepackage{algorithmicx}
\usepackage{algpseudocode}
\usepackage{balance}
\definecolor{mycitecolor}{RGB}{0, 102, 204}  

\usepackage{bm}
\makeatletter
\hypersetup{
  colorlinks=true,
  citecolor=mycitecolor,
  linkcolor=black,
  urlcolor=blue
}
\usepackage[table]{xcolor}
\makeatother

\newcommand{\model}{TIPS}
\newcommand{\teacher}{Bias Teacher Ensemble}

\copyrightyear{2026}
\acmYear{2026}
\setcopyright{cc}
\setcctype{by}
\acmConference[KDD 2026] {Proceedings of the 32nd ACM SIGKDD Conference on Knowledge Discovery and Data Mining V.2}{August 9--13, 2026}{Jeju Island, Republic of Korea.}
\acmBooktitle{Proceedings of the 32nd ACM SIGKDD Conference on Knowledge Discovery and Data Mining V.2 (KDD 2026), August 9--13, 2026, Jeju Island, Republic of Korea}
\acmISBN{979-8-4007-2259-2/2026/08}
\acmDOI{10.1145/3770855.3817749}

\begin{document}

\title{Integrating Inductive Biases in Transformers via Distillation for Financial Time Series Forecasting}


\definecolor{darkgreen}{rgb}{0.00, 0.40, 0.00}



\author{Yu-Chen Den}\authornote{Equal contribution}
\affiliation{%
  \institution{SinoPac Holdings}
  \city{Taipei}
  \country{Taiwan}
}
\email{abnerden@sinopac.com}

\author{Kuan-Yu Chen}\authornotemark[1]
\affiliation{%
  \institution{SinoPac Holdings}
  \city{Taipei}
  \country{Taiwan}
}
\email{lavamore@sinopac.com}

\author{Kendro Vincent}\authornote{Corresponding author}
\affiliation{%
  \institution{National Chengchi University}
  \city{Taipei}
  \country{Taiwan}
}
\email{kendrov@nccu.edu.tw}

\author{Tien-Hao Chang}
\affiliation{%
  \institution{SinoPac Holdings}
  \city{Taipei}
  \country{Taiwan}
}
\email{darby@sinopac.com}

\renewcommand{\shortauthors}{Yu-Chen Den, Kuan-Yu Chen, Kendro Vincent, \& Darby Tien-Hao Chang}

\begin{abstract}
Transformer-based models have been widely adopted for generic time-series forecasting due to their high representational capacity and architectural
flexibility.
However, many Transformer variants implicitly assume stationarity and stable temporal dynamics—assumptions that are routinely violated in financial markets characterized by regime shifts and non-stationarity.
Empirically, state-of-the-art time-series Transformers often underperform even vanilla Transformers on financial tasks, while simpler architectures with distinct inductive biases, such as CNNs and RNNs, can achieve stronger performance with substantially lower complexity.
At the same time, no single inductive bias dominates across markets or regimes, suggesting that robust financial forecasting requires integrating complementary temporal priors.

We propose \textbf{TIPS} (\textbf{T}ransformer with \textbf{I}nductive \textbf{P}rior \textbf{S}ynthesis), a knowledge distillation framework that synthesizes diverse inductive biases—causality, locality, and periodicity—within a unified Transformer.
TIPS first trains bias-specialized Transformer teachers via attention masking, 
then distills their collective knowledge into a single student model that exhibits regime-dependent alignment with different inductive biases.
Across four major equity markets, TIPS achieves state-of-the-art performance, outperforming strong ensemble baselines by 55\%, 9\%, and 16\% in annual return, Sharpe ratio, and Calmar ratio, respectively, while requiring only 38\% of the inference-time computation.
Further analyses show that TIPS generates statistically significant excess returns beyond both vanilla Transformers and its teacher ensembles, and exhibits regime-dependent behavioral alignment with classical architectures during their profitable periods.
These results highlight the importance of regime-dependent
inductive bias utilization for robust generalization in non-stationary financial time series.
\end{abstract}

\ccsdesc[500]{Information systems~Data mining}
\ccsdesc[500]{Applied computing~Economics}
\ccsdesc[500]{Computing methodologies~Machine learning}

\keywords{Financial Time Series Forecasting; Inductive Bias; 
Transformer; Knowledge Distillation; Attention Mechanism}



\maketitle

\input{content/intro}

\input{content/preliminary}
\input{content/method}
\input{content/expr}
\input{content/discussion}
\input{content/related}
\input{content/conclude}

\bibliographystyle{ACM-Reference-Format}
\balance
\bibliography{main}

\appendix
\input{content/appendix}
\end{document}

%% file: content/intro.tex
\begin{figure}[H]
    \centering
    \includegraphics[width=0.9\columnwidth]{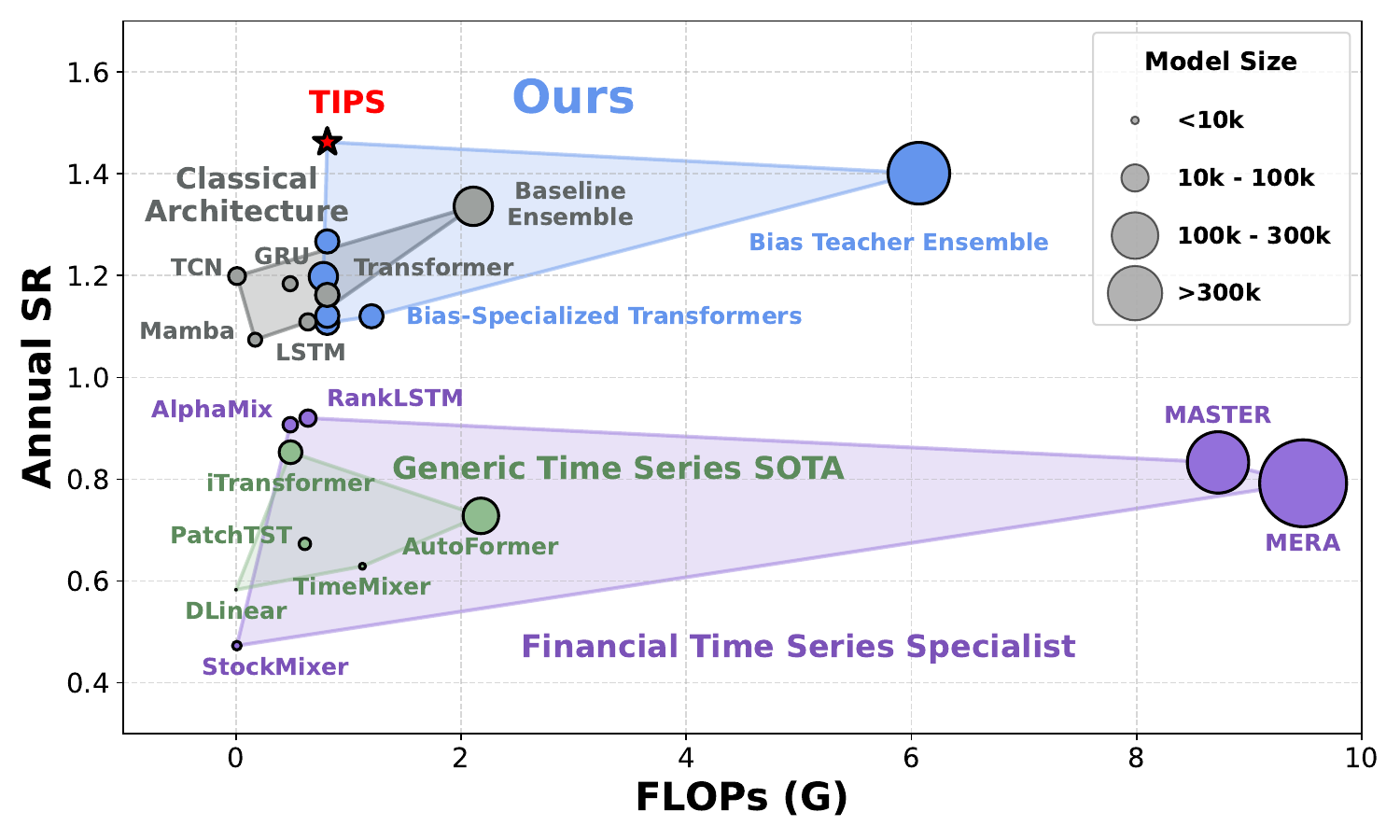}
    \caption{Performance–efficiency trade-off across generic time-series models, financial forecasting models, and classical architectures evaluated across multiple equity markets. The figure highlights substantial variation in performance and computational cost across model families, with TIPS achieving the strongest performance among the evaluated methods while maintaining low inference-time overhead.
    }
    \label{fig:teaser}
\end{figure}

\section{Introduction}
\label{sec:intro}

Transformer-based architectures~\citep{vaswani2017attention} have become a dominant modeling paradigm across domains such as natural language processing, computer vision, and audio processing, owing to their strong representational capacity and minimal structural assumptions~\citep{brown2020language,devlin2019bert,liu2021swin,baevski2020wav2vec}.
By relying on data-driven self-attention to capture complex dependencies, Transformers are widely viewed as flexible and broadly applicable backbones.
Motivated by this success, recent work has increasingly adopted Transformers for time series forecasting~\citep{wu2021autoformer,nie2022time,liu2023itransformer}, including applications to financial markets that require modeling both temporal dynamics and cross-sectional interactions~\citep{li2024master,sun2023mastering,yu2024miga,liu2025mera}.

Despite the generality of Transformers, modeling assumptions effective for generic time series forecasting often fail to transfer to financial settings.
While many generic time series Transformers are designed to exploit relatively stable patterns such as seasonality or long-term periodicity, financial markets exhibit pronounced non-stationarity and regime-dependent behavior~\citep{azariadis1998financial,hu2025fintsb}.
On the other hand, although financial forecasting models often incorporate domain-specific structural assumptions about temporal dependencies, these assumptions are typically fixed at the model architectural level.
Existing approaches primarily address regime variation through input conditioning or output-level expert routing~\citep{sun2023mastering,liu2025mera}, rather than adapting the underlying assumptions, potentially limiting their robustness under regime shifts.

As illustrated in Figure~\ref{fig:teaser}, state-of-the-art models developed for general time series benchmarks frequently underperform even vanilla Transformers when applied to financial data, and a similar phenomenon is observed for specialized financial forecasting models.
Moreover, lightweight architectures such as GRU, LSTM, Mamba, and TCN often surpass high-capacity Transformers while using orders of magnitude fewer parameters and computation~\citep{hochreiter1997long,chung2014empirical,lea2017temporal,gu2024mamba}.
Given that these simpler architectures outperform substantially larger models, the performance gap cannot be attributed to model capacity alone.
Instead, it suggests that financial forecasting performance is governed by the structural constraints that determine how historical information is utilized—specifically, which temporal dependencies (e.g., short-term locality, long-range structure, or causal ordering) are emphasized.
We refer to such structural preferences as the model’s \emph{inductive bias}.

The importance of inductive bias is further underscored by cross-market heterogeneity: no single architecture consistently dominates across markets (\cref{tab:main_results}), as different regimes favor distinct temporal priors.
This raises a central challenge: \emph{how can we integrate complementary inductive biases into an unified model that adapts to changing market conditions?}
Transformers provide a natural substrate for addressing this challenge, as their attention mechanism enables diverse structural priors to be encoded through masking and input design~\citep{press2021train,nie2022time,sun2025penguin} while preserving architectural homogeneity.
However, naively combining multiple inductive biases within a single model often leads to performance degradation, a phenomenon we refer to as the \emph{merging penalty}.

To address this problem, we propose \textbf{TIPS} (\textbf{T}ransformer with \textbf{I}nductive \textbf{P}rior
\textbf{S}ynthesis), a knowledge-distillation framework that synthesizes diverse inductive biases into a single Transformer.
TIPS operates in two stages. First, we train multiple bias-specialized Transformer teachers that encode distinct priors—such as causality, locality, and periodicity—via attention masking and input patching.
Second, we distill their collective knowledge into a compact student using aggressive regularization that prevents rigid teacher mimicry while enabling flexible synthesis.
As a result, the student exhibits regime-dependent utilization of inductive biases, achieving ensemble-level robustness with the inference cost of a single model.
\textbf{Our contributions are threefold:}

\begin{itemize}[leftmargin=*, itemsep=2pt, parsep=0pt]
    \item \textbf{Systematic Analysis of Inductive Biases.}
    We conduct a comprehensive empirical study demonstrating that different market regimes favor distinct inductive biases. Through controlled experiments within a unified Transformer backbone, we further show that naively merging multiple biases in a single model degrades performance,
    highlighting the limitations of joint multi-bias training.

    \item \textbf{Efficient Inductive Prior Synthesis.}
    We introduce TIPS, a distillation-based framework that integrates diverse inductive biases into a single Transformer.
    TIPS achieves state-of-the-art performance across four major equity markets, outperforming the strongest non-proposed ensemble baselines by up to 55\% in annual return, 9\% in Sharpe ratio, and 16\% in Calmar ratio, while requiring substantially lower inference-time computation.
    \item \textbf{Evidence of Conditional Bias Activation.}
    Through multi-level empirical and behavioral analyses, we provide evidence that TIPS exhibits regime-dependent alignment with different inductive biases, aligning with specific architectural behaviors during their profitable periods rather than uniformly replicating any single model. These findings help explain TIPS’s strong generalization under non-stationary financial dynamics.
\end{itemize}

%% file: content/preliminary.tex
\section{Preliminary: Financial Time Series Forecasting for Portfolio Construction}
\label{sec:preliminary}

\subsection{Task Formalization and Domain Challenges}
\label{sec:challenges}

\paragraph{\textbf{Problem Formulation.}}
We formulate financial forecasting as a ranking task over a universe of $S$ stocks observed across $T$ trading days, each described by $F$ features.
The input data are represented as $\bm{X} \in \mathbb{R}^{S \times T \times F}$.
Our objective is to learn a model $f_{\theta}$ that maps historical observations to a predicted cross-sectional ranking, optimized via a ranking-based loss:
$\min_{\theta} \ \mathcal{L}(\bm{y}, f_{\theta}(\bm{X}))$, 
where $\bm{y} \in \mathbb{R}^{S}$ is typically defined by the magnitude of each stock’s $q$-day future return and used as the continuous target ranking score.

At inference time, for a trading day \(t\), we construct a long-only portfolio by selecting the top-$k$ ranked stocks and assigning portfolio weights $\bm{w}_t \in \mathbb{R}^{k}$.
Performance is evaluated using standard risk-adjusted metrics, including the Sharpe Ratio and Calmar Ratio, computed from daily portfolio returns $\bm{p}_t = \bm{w}_t^{\intercal} \bm{C}_t$, where $\bm{C}_t \in \mathbb{R}^{k \times q}$ denotes the realized returns at trading day $t$.

\paragraph{\textbf{Domain Challenges.}}
Financial time series forecasting differs from standard temporal modeling primarily due to the instability of its underlying market dynamics, which evolve in response to economic cycles, policy interventions, and behavioral feedback, resulting in frequent regime shifts and distributional drift~\citep{hamilton1989new,andreou2002detecting,xu2024rhine}.
Moreover, predictive signals in financial data are weak and highly context-dependent, with structure that varies across time and markets~\citep{merton1980estimating, forbes2002no,lehkonen2014timescale}.
Modeling assumptions that are beneficial under one regime may become harmful under another, leading to performance degradation in out-of-sample performance.
These characteristics imply that robust financial forecasting requires models whose inductive assumptions can adapt to changing market conditions, rather than relying on a fixed structural prior.

\subsection{Inductive Biases and the Merging Penalty}
\label{sec:prelim-inductive-bias}

Building on the need for adaptive modeling assumptions discussed in \cref{sec:challenges}, we examine inductive biases reflected in the architectures that perform well in financial forecasting (e.g., CNNs, RNNs, and state-space models; see Figure~\ref{fig:teaser}).
In particular, we focus on \textbf{causality}, \textbf{locality}, and \textbf{periodicity}, which capture complementary aspects of market dynamics related to sequential dependence, noise suppression, and recurring temporal structure.
Table~\ref{tab:bias_comparison} summarizes how these biases are implicitly encoded in classical architectures and how they can be explicitly instantiated in Transformer models through attention masking and positional design.

\begin{itemize}[leftmargin=*, itemsep=2pt, parsep=0pt]

\item \textbf{Causality.}
Causality enforces unidirectional dependence on past observations ($\leq t$), preventing look-ahead bias while supporting momentum and trend-following behavior.
It is intrinsic to recurrent and state-space models and can be imposed in Transformers via causal attention masking.
In practice, attention-based models without explicit causal constraints allow earlier time steps to attend to later observations within the same historical window, which can hinder generalization and degrade out-of-sample performance.

\item \textbf{Locality.}
Locality prioritizes recent observations and attenuates distant noise, which is particularly important in low signal-to-noise financial environments.
While CNN-based architectures encode locality through finite receptive fields,  
Transformers incorporate locality through mechanisms such as temporal patching~\citep{nie2022time} or distance-aware attention biases~\citep{press2021train}.
Empirically, without locality constraints, long-range attention may be emphasized on noise-dominated dependencies.

\item \textbf{Periodicity.}
Periodicity captures recurring temporal patterns such as calendar effects and institutional trading cycles.
Dilated CNNs and recurrent models represent periodic structure implicitly through long-range states, whereas Transformers encode it via fixed or learnable positional biases~\citep{raffel2020exploring,sun2025penguin}.
This bias is particularly effective in markets exhibiting persistent seasonal or cyclical behavior, but its utility may weakened under regime shifts.
\end{itemize}

\input{tables/inductive-bias.tex}

\input{tables/merge-penalty}

\paragraph{\textbf{Empirical Analysis: The Merging Penalty.}}
To examine how multiple inductive biases interact within a single model, we compare three strategies: \emph{single-bias specialization}, \emph{joint multi-bias training}, and \emph{ensemble inference}.
As shown in Table~\ref{tab:merge_penalty}, different biases dominate under different market conditions (e.g., causality on NI225, periodicity on CSI300), highlighting their complementary strengths.

However, naively combining multiple biases in a single model consistently underperforms the strongest single-bias model, a phenomenon we term the \emph{merging penalty}. This suggests that one shared set of parameters struggles to fully commit to very different inductive assumptions, and instead converging to a solution that is suboptimal across regimes.
Ensemble methods mitigate this issue by training separate models for each bias, which leads to better performance but also higher inference cost.
Together, these observations reveal a clear design challenge: retaining the benefits of bias specialization without incurring ensemble overhead.

%% file: tables/inductive-bias.tex
\begin{table}[t]
    \centering
    \caption{Inductive biases in financial time series and their instantiation in classical models and Transformers. Global context denotes unconstrained attention.}
    \label{tab:bias_comparison}
    \resizebox{1.0\linewidth}{!}{%
    \begin{tabular}{lcc}
        \toprule
        \textbf{Inductive Bias} & \textbf{Classical Models} & \textbf{Transformer Design} \\
        \midrule
        Global Context & Standard Transformer & Vanilla Attention \\
        Causality & RNNs, Mamba & Causal Attention Masks \\
        Locality & TCN & Input Patching, Recency Decay Bias \\
        Periodicity & Dilated CNN, LSTM Memory & Positional Bias (fixed/learnable) \\
        \bottomrule
    \end{tabular}
    }
\end{table}

%% file: tables/merge-penalty.tex
\begin{table}[t]
    \centering
    \caption{Performance (Annual Sharpe Ratio) of different bias integration strategies. \textbf{Bold}: best per market; $\dagger$: outperforms Vanilla Transformer 
    (row 1). Higher is better.}
    \label{tab:merge_penalty}
    \resizebox{0.95\columnwidth}{!}{%
    \begin{tabular}{lcccc|c}
    \toprule
        \textbf{Method} & \textbf{CSI300} & \textbf{CSI500} & \textbf{NI225} & \textbf{SP500} & \textbf{Avg.} \\
        \midrule
        Vanilla Transformer & 0.915 & \textbf{1.662} & 0.774 & 1.297 & 1.162 \\
        \midrule
        \multicolumn{6}{l}{\textit{Single Inductive Bias}} \\
        \quad + Causality & 1.032$^\dagger$ & 1.510 & \textbf{0.841}$^\dagger$ & 1.261 & 1.161 \\
        \quad + Locality & 0.972$^\dagger$ & 1.477 & 0.695 & 1.283 & 1.107 \\
        \quad + Periodicity & 1.056$^\dagger$ & 1.496 & 0.693 & 1.234 & 1.120 \\
        \midrule
        \multicolumn{6}{l}{\textit{Multiple Biases (Merged)}} \\
        \quad Causality + Locality & 1.023$^\dagger$ & 1.502 & 0.781$^\dagger$ & 1.288$^\dagger$ & 1.149 \\
        \quad Causality + Periodicity & 0.927$^\dagger$ & 1.520 & 0.811$^\dagger$ & 1.370$^\dagger$ & 1.157 \\
        \midrule
        \multicolumn{6}{l}{\textit{Multiple Biases (Ensemble)}} \\
        \quad Causality + Locality & \textbf{1.115}$^\dagger$ & 1.612 & 0.747 & \textbf{1.442}$^\dagger$ & \textbf{1.229}$^\dagger$ \\
        \quad Causality + Periodicity & 1.125$^\dagger$ & 1.598 & 0.773 & 1.410$^\dagger$ & 1.227$^\dagger$ \\
    \bottomrule
    \end{tabular}}
\end{table}

%% file: content/method.tex
\section{TIPS: Transformer with Inductive Prior Synthesis}
\label{sec:method}

\begin{figure*}[t]
    \centering    \includegraphics[width=0.8\textwidth]{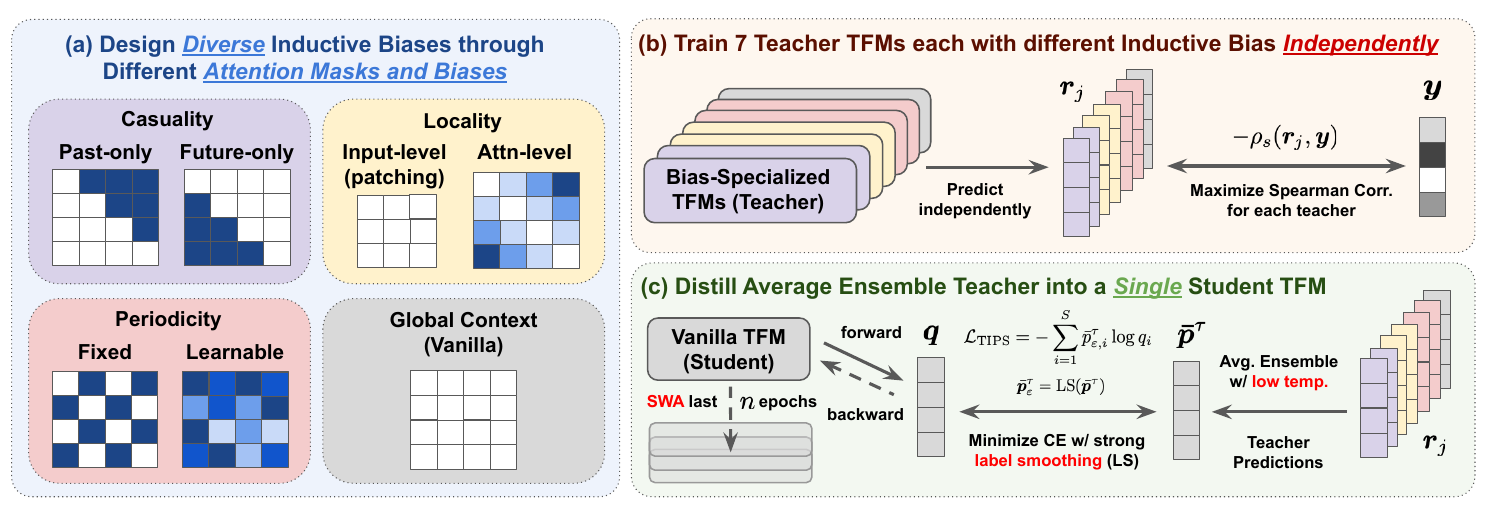}
    \caption{Overview of the TIPS training framework. (a) Bias-specialized Transformer (TFM) teachers are constructed via different attention masks or positional biases (Colors indicate where the masks and biases are applied). (b) Teachers are trained independently for ranking prediction. (c) Teacher predictions are averaged and distilled into a single student model.}
    \label{fig:training-framework}
\end{figure*}

We introduce \textbf{TIPS} (\textbf{T}ransformer with \textbf{I}nductive \textbf{P}rior \textbf{S}ynthesis), a knowledge distillation framework that integrates diverse specialized temporal biases into a single Transformer.
TIPS operates in two stages. First, we train a set of bias-specialized Transformer teachers that independently encode distinct temporal priors through attention masking, thereby isolating bias-specific signals (\cref{sec:teacher_models}).
Second, we distill their collective knowledge into a unified student model using an aggressive regularization strategy that encourages consensus learning rather than rigid teacher mimicry (\cref{sec:distillation}).
An overview of the TIPS training framework is illustrated in \cref{fig:training-framework}.

\subsection{Teacher Models: Specializing Priors via Attention Masking}
\label{sec:teacher_models}

Building on the inductive biases identified in \cref{sec:prelim-inductive-bias}, we instantiate each bias through localized modifications to the attention mechanism rather than architectural changes.
This design maintains structural homogeneity across the teacher set, ensuring that performance differences arise solely from distinct temporal modeling assumptions rather than variations in model capacity or depth.

\paragraph{\textbf{Attention with Structural Masks and Biases.}}
We extend the standard Transformer attention mechanism to explicitly encode structural priors.
Given query, key, and value matrices $\bm{Q}, \bm{K}, \bm{V} \in \mathbb{R}^{T \times d}$, attention is computed as
\begin{equation}
\text{Attention}(\bm{Q}, \bm{K}, \bm{V}; \bm{M}, \bm{B})
=
\text{softmax}\!\left(
\frac{\bm{Q}\bm{K}^\top}{\sqrt{d}} + \bm{M} + \bm{B}
\right)\bm{V},
\label{eq:attention}
\end{equation}
where $\bm{M} \in \{0, -\infty\}^{T \times T}$ denotes a structural attention mask and $\bm{B} \in \mathbb{R}^{T \times T}$ is an additive attention bias.
Both $\bm{M}$ and $\bm{B}$ are shared across all $L$ Transformer layers to enforce consistent temporal priors throughout the network.
Vanilla Transformers recover unconstrained bidirectional attention by setting $\bm{M} = \bm{B} = \bm{0}$.
We note that some locality mechanisms (e.g., patching) operate at the input level rather than through explicit attention modification.

\paragraph{\textbf{Causality: Directional Decomposition.}}
To isolate directional temporal dependencies, we decompose bidirectional attention into two complementary unidirectional patterns:
\begin{itemize}[leftmargin=*, itemsep=2pt, parsep=0pt]
    \item \textbf{Past-only Mask} ($\mathcal{T}_{\text{past}}$).
    Standard causal masking where
    $M_{ij}=0$ if $i \ge j$ and $M_{ij}=-\infty$ otherwise.
    This enforces left-to-right attention over historical observations and captures momentum or trend-following effects.

    \item \textbf{Future-only Mask} (Reverse-direction Mask, $\mathcal{T}_{\text{future}}$).
    Reverse causal masking where
    $M_{ij}=0$ if $i < j$ and $M_{ij}=-\infty$ otherwise.
    Importantly, this mask operates \emph{only within the same historical window} as the past-only mask and does \emph{not} expose any post-decision information.
    It can be viewed as inducing a backward-directed attention pattern over the historical window, serving as an inductive prior over reverse temporal dependencies rather than a form of look-ahead.
\end{itemize}

\paragraph{\textbf{Locality: Structural Aggregation and Distance Decay.}}
We encode locality through two complementary mechanisms:
\begin{itemize}[leftmargin=*, itemsep=2pt, parsep=0pt]
    \item \textbf{Patching} ($\mathcal{T}_{\text{patch}}$).
    Following~\citet{nie2022time}, we partition the input
    $\bm{X} \in \mathbb{R}^{S \times T \times F}$
    into overlapping temporal patches of length $p$ and stride $s$
    using an \texttt{Unfold} operation.
    Each patch is projected into a single token via
    \begin{equation}
        \bm{X}^{\text{patch}}
        =
        \texttt{Unfold}(\bm{X}, p, s)\,\bm{W}_{\text{proj}}
        \in \mathbb{R}^{S \times T' \times d},
    \end{equation}
    where
    $T' = \lfloor (T - p)/s \rfloor + 1$.
    Cross-patch attention then models interactions between short-term temporal summaries.

    \item \textbf{ALiBi} ($\mathcal{T}_{\text{ALiBi}}$).
    To preserve fine-grained temporal resolution, we apply a distance-based decay bias
    $B_{ij}^{h} = -m_h |i - j|$
    following~\citet{press2021train}.
    Each attention head $h$ uses a slope
    $m_h = 2^{-8/h}$,
    encouraging attention to focus on local context while retaining access to long-range dependencies.
\end{itemize}

\paragraph{\textbf{Periodicity: Fixed and Learned Recurrence.}}
To capture recurring market structures, we employ both fixed and learnable periodic biases:
\begin{itemize}[leftmargin=*, itemsep=2pt, parsep=0pt]
    \item \textbf{Fixed Periodic Bias} ($\mathcal{T}_{\text{fixed}}$).
    Inspired by dilated convolutions, each attention head $h$ is assigned a predefined period $p_h$
    (e.g., weekly or monthly).
    Following~\citet{sun2025penguin}, the bias is defined as
    \begin{equation}
        B_{ij}^{\text{fixed},h} =
        \begin{cases}
        \beta_{ij}, & \beta_{ij} < p_h/2, \\
        p_h - \beta_{ij}, & \text{otherwise},
        \end{cases}
    \end{equation}
    where
    $\beta_{ij} = |i - j| \bmod p_h$.

    \item \textbf{Learnable Relative Bias} ($\mathcal{T}_{\text{learn}}$).
    We adopt Relative Positional Bias (RPB)~\citep{raffel2020exploring}, defined as a learnable function of relative index offsets:
    \begin{equation}
        B_{ij}^{\text{learn}} = \text{RPB}(i - j;\theta_{\text{RPB}}),
    \end{equation}
    enabling the model to capture distance- and recurrence-dependent temporal patterns without assuming a fixed period.
\end{itemize}

\paragraph{\textbf{Teacher Training.}}
Each teacher $\mathcal{T}$ shares the same $L$-layer Transformer backbone with hidden dimension $d$ and $H$ attention heads, differing only in their structural masks and biases.
Given an input $\bm{X}$, the teacher produces ranking logits $\bm{r} \in \mathbb{R}^{S}$ and is optimized using differentiable Spearman correlation~\citep{blondel2020fast}:
\begin{equation}
\mathcal{L}_{\text{teacher}} = -\rho_s(\bm{r}, \bm{y}),
\end{equation}
where $\bm{y}$ denotes the ground-truth ranking. This objective directly aligns with portfolio construction and is applied uniformly across all teachers to ensure fair comparison.

\paragraph{\textbf{Vanilla Transformer as Seventh Teacher.}}
In addition to six bias-specialized teachers, we include a vanilla Transformer with $\bm{M}=\bm{B}=\bm{0}$ as the seventh teacher $\mathcal{T}_{\text{vanilla}}$.
Including the vanilla model ensures that the distillation process has access to unconstrained bidirectional attention patterns, providing a strong and widely used baseline within the teacher ensemble. The complete teacher set is therefore
\begin{equation}
\boldsymbol{\mathcal{T}}=\{
\mathcal{T}_{\text{past}},\mathcal{T}_{\text{future}},
\mathcal{T}_{\text{patch}},\mathcal{T}_{\text{ALiBi}},
\mathcal{T}_{\text{fixed}},\mathcal{T}_{\text{learn}},
\mathcal{T}_{\text{vanilla}}
\},
\end{equation}
which we refer to as the \textbf{\teacher{}}.

\subsection{Knowledge Distillation Framework}
\label{sec:distillation}

Having trained seven bias-specialized teachers independently, we now synthesize their knowledge into a unified student model.
The central challenge is to integrate heterogeneous inductive priors without forcing the student to rigidly imitate any individual teacher.
To address this, TIPS employs an aggressively regularized distillation framework that encourages consensus learning while preserving fine-grained ranking signals.

\paragraph{\textbf{Distillation Objective.}}
The student $\mathcal{S}_{\phi}$ adopts the same Transformer backbone as the teachers, but uses unconstrained bidirectional attention($\bm{M}=\bm{B}=\bm{0}$), allowing it to flexibly combine learned priors.
Given an input $\bm{X}$, we first construct a soft ensemble target by averaging teacher logits from the \teacher{} and applying temperature scaling:
\begin{equation}
\bar{\bm{p}}^{\tau}
=
\text{softmax}\!\left(
\frac{1}{\tau}
\cdot
\frac{1}{|\boldsymbol{\mathcal{T}}|}
\sum_{j=1}^{|\boldsymbol{\mathcal{T}}|}
\bm{r}_j
\right)
\in \mathbb{R}^{S},
\label{eq:ensemble}
\end{equation}
where
$\bm{r}_j = \mathcal{T}_j(\bm{X}) \in \mathbb{R}^{S}$ denotes the ranking logits produced by the $j$-th teacher and $|\boldsymbol{\mathcal{T}}| = 7$. To reduce overfitting to specific teacher outputs and improve generalization, we apply label smoothing to the ensemble distribution:
\begin{equation}
\bar{\bm{p}}^{\tau}_{\varepsilon}
=
(1-\varepsilon)\,\bar{\bm{p}}^{\tau}
+
\varepsilon\,\frac{1}{S}\bm{1},
\label{eq:label_smooth}
\end{equation}
where $\varepsilon$ is the smoothing coefficient and $\bm{1} \in \mathbb{R}^{S}$ denotes the all-ones vector .
The student prediction $\bm{q} = \text{softmax}(\mathcal{S}_{\phi}(\bm{X}))$
is trained by minimizing the cross-entropy loss
\begin{equation}
\mathcal{L}_{\text{TIPS}}(\phi) =
-\sum_{i=1}^{S}
\bar{p}^{\tau}_{\varepsilon,i}
\log q_i.
\end{equation}
We further apply Stochastic Weight Averaging (SWA)~\citep{izmailov2018averaging}
during the final $n$ epochs of training:
\begin{equation}
\phi_{\text{SWA}}^{(t)}
=
\frac{1}{n}
\sum_{i=0}^{n-1}
\phi^{(t-i)},
\label{eq:swa}
\end{equation}
where $\phi^{(t)}$ denotes the student parameters at epoch $t$. SWA acts as an additional regularization mechanism by favoring flatter minima.
The student is trained \emph{exclusively} using teacher predictions, without direct supervision from ground-truth labels.
Including a vanilla Transformer teacher ($\mathcal{T}_{\text{vanilla}}$) anchors the distilled target to unconstrained bidirectional attention, providing a stable baseline when specialized inductive priors are not beneficial.

\paragraph{\textbf{Mechanism for Robust Synthesis.}}
The components introduced above each target a distinct aspect 
of integrating heterogeneous inductive priors:
\begin{itemize}[leftmargin=*, itemsep=2pt, parsep=0pt]
  \item \textbf{Sharp Ranking Targets:} low-temperature distillation
  preserves fine-grained ranking signals.
  \item \textbf{Calibration-Aware Smoothing:} aggressive label smoothing
  mitigates overfitting to individual teacher behaviors.
  \item \textbf{Stability Across Regimes:} stochastic weight averaging
  biases optimization toward flatter, more robust solutions.
\end{itemize}
Together, these components enable the student to synthesize complementary inductive biases without rigidly imitating any individual teacher.

%% file: content/expr.tex
\section{Experiments}
\label{sec:experiments}

\input{tables/dataset-info}

\paragraph{\textbf{Datasets with Diverse Market Regimes.}}
We evaluate
\model{}
across four major equity markets: CSI300, CSI500, NI225, and SP500. For each market, we construct datasets using eight temporal features, including OHLCV and moving averages, following prior works~\citep{feng2019temporal,yoo2021accurate,sun2023mastering}.
Detailed dataset statistics are summarized in Table~\ref{tab:dataset_stats}, while exact feature definitions and label formulations are provided in~\cref{subsec:features}.
The test period from 2021 to 2024 spans multiple market regimes, including bull markets, bear markets, and high-volatility phases driven by regulatory changes, monetary policy shifts, and inflation surges. This setting induces pronounced non-stationarity and provides a challenging testbed for model robustness (\cref{subsec:market_regime}). 
Across markets, return characteristics and temporal dependencies vary substantially: buy-and-hold Sharpe ratios range from 0.265 to 0.524, and lag-1 weekly return autocorrelations
differ in both magnitude and sign, indicating heterogeneous short-horizon serial dependence that may favor distinct inductive biases in different markets, such as momentum-driven behavior in some markets and mean-reversion in others.

\input{tables/main-perf}

\paragraph{\textbf{Baselines and Evaluation Setup.}}
To comprehensively evaluate the effectiveness of \model{}, we compare it against three categories of representative baselines:
\textbf{(1) Generic time series SOTA} includes DLinear~\citep{zeng2023transformers}, TimeMixer~\citep{wangtimemixer}, AutoFormer~\citep{wu2021autoformer}, PatchTST~\citep{nie2022time}, and iTransformer~\citep{liu2023itransformer};
\textbf{(2) Classical architectures} includes TCN~\citep{lea2017temporal}, GRU~\citep{chung2014empirical}, LSTM~\citep{hochreiter1997long}, Transformer~\citep{vaswani2017attention}, and Mamba~\citep{gu2024mamba}; and
\textbf{(3) Financial time series specialists} includes RankLSTM~\citep{feng2019temporal}, AlphaMix~\citep{sun2023mastering}, MASTER~\citep{li2024master}, StockMixer~\citep{fan2024stockmixer}, and MERA~\citep{liu2025mera}.
For each category, we additionally report an \textbf{Ensemble} baseline, computed by averaging pre-softmax logits across constituent models within that category, followed by softmax to obtain final ranking scores and portfolio weights.
Each constituent model is trained with five random seeds; Ensemble results are aggregated accordingly. Performance is evaluated using portfolio-based metrics, including Annualized Return (AR), Sharpe Ratio (SR), and Calmar Ratio (CR), averaged across seeds.  Hyperparameter settings and implementation details for both TIPS and baselines are provided in \cref{subsec:baseline,subsec:implement-detail}, with the full evaluation protocol is described in \cref{subsec:eval-method}.

\subsection{Main Results}
\label{sec:main_results}

\cref{tab:main_results} compares \model{} with state-of-the-art baselines across four major equity markets.
Overall, the results reveal three key findings concerning generalization performance, inductive bias modeling, and inference efficiency.

\paragraph{\textbf{Superior Generalization via Bias Synthesis.}}
\model{} achieves the strongest overall performance across all markets, with an average SR of 1.454 and an average AR of 0.907.
Compared with the strongest ensemble baseline outside our masking framework (Classical Architecture Ensemble), \model{} improves AR by 54.8\% and SR by 8.8\%.
It also consistently outperforms the best generic time series SOTA model in terms of all metrics.
These results indicate that synthesizing multiple temporal inductive biases within a unified framework yields more robust generalization under non-stationary market conditions than relying on a single architectural prior.

\paragraph{\textbf{Sufficiency of Attention Masking for Bias Encoding.}}
The \teacher{} demonstrates that diverse inductive biases can be effectively encoded through attention masking alone, without introducing architectural heterogeneity.
Specifically, it outperforms ensembles of classical architectures (average AR: 0.703 vs.\ 0.586) as well as ensembles of generic time series SOTA models (average AR: 0.278). This finding suggests that structural priors such as causality and locality can be explicitly isolated and exploited within a homogeneous Transformer backbone, rather than requiring separate model architectures for each bias.

\paragraph{\textbf{Distillation as an Effective Knowledge Integrator.}}
Beyond ensembling, the proposed distillation strategy further consolidates cross-bias knowledge into a single model.
\textsc{TIPS} improves upon the Bias Teacher Ensemble by 29.0\% in AR (0.907 vs.\ 0.703), exhibiting a clear student-surpasses-teacher effect.
Importantly, this gain is achieved with a $7\times$ reduction in inference time computation, enabling ensemble-level robustness at the cost of a single Transformer model.
Together, these results confirm that distillation provides an effective mechanism for integrating complementary inductive biases while maintaining high inference efficiency. We further verify in \cref{subsec:scale_analysis} that this gain 
cannot be replicated by simply scaling up vanilla Transformers.

\subsection{Ablation Studies}
\label{sec:ablation}

\paragraph{\textbf{Sufficiency of Teacher-only Supervision.}}
\cref{tab:ablation_supervision} compares mixed supervision strategies against pure distillation, where the loss is a weighted combination of the distillation term and the ground-truth ranking target using Spearman correlation.
Adding ground-truth supervision does not yield consistent gains over pure distillation ($1.0 \times \text{distill}$) across markets, suggesting that teacher predictions already provide sufficient supervisory signal in our setting.
We therefore adopt pure distillation as the default, which also avoids introducing an additional loss-weight hyperparameter.

\paragraph{\textbf{Synergistic Effects of Distillation Regularization.}}
\cref{tab:ablation_reg} evaluates the contribution of each regularization component used in \model{} by incrementally adding them to vanilla distillation.
Vanilla distillation alone does not outperform the Bias Teacher Ensemble, indicating that naive teacher imitation is insufficient.
Introducing low-temperature distillation improves test SR by preserving fine-grained ranking signals, but widens the train-test gap, reflecting higher overfitting risk under non-stationary markets.
Aggressive label smoothing mitigates overfitting to individual teacher logits, reducing the gap while further improving test performance.
Applying SWA yields the strongest and most consistent gains across markets, achieving the best overall SR while further closing the train-test gap.
These results confirm that the proposed regularization components act synergistically: each stage improves not only test performance but also generalization, and all components are necessary for effective inductive bias synthesis.
Among these, aggressive label smoothing warrants closer inspection: by redistributing probability mass toward a uniform distribution, it may inadvertently suppress the student's ability to make decisive stock selections.
\input{tables/ablation-supervision}
\input{tables/ablation-distill}
\input{tables/score-concentration}

\paragraph{\textbf{Label Smoothing Regularizes without Suppressing Decisive Signals.}}
To further characterize the effect of label smoothing, we measure the average portfolio weight assigned to the top-1 selected stock as a proxy for extreme-score concentration. As shown in \cref{tab:score_concentration}, removing label smoothing substantially increases top-1 concentration across all markets, whereas \model{} and \model{} w/o SWA remain nearly identical.
This confirms that label smoothing regularizes excessive concentration without eliminating decisive signals: the final model remains more concentrated than all individual teacher groups, indicating that the student retains the ability to make confident stock selections.

\paragraph{\textbf{From Teacher Mimicry to Consensus Synthesis.}}
Table~\ref{tab:similarity_master} reports teacher--student rank similarity across regularization stages, which is measured as the average Spearman correlation between predictions of the seven teachers and the distilled student. 
As regularization is strengthened, both the average similarity (–31.8\%) and the inter-teacher variance decrease (from 0.179 to 0.097).
High variance under vanilla distillation indicates unstable alignment with conflicting teacher behaviors.
In contrast, the reduced variance under the full TIPS framework reflects convergence toward a stable, consensus-oriented representation.
This observation clarifies how the distilled student can outperform ensemble inference despite lower average similarity to individual teachers.

\paragraph{\textbf{Robustness Across Bias Configurations.}}
Table~\ref{tab:ablation_bias} studies distillation performance under different bias configurations.
Distillation consistently outperforms ensemble inference even when only a single type of bias is present (e.g., causality only: Avg.\ SR 1.451 vs.\ 1.408), and the performance gap increases as more biases are combined.
The largest gains are observed when all biases are included, highlighting the benefit of increased teacher diversity for distillation.
Performance improvements are more pronounced in volatile markets such as CSI300 (+16.6\%), where frequent regime shifts favor consensus-based representations.
In more stable markets such as SP500, the smaller gap suggests that ensemble inference and distilled synthesis provide comparable benefits.

\input{tables/student-teacher-similarity}
\input{tables/ablation-mask}

%% file: tables/dataset-info.tex
\begin{table}[t]
  \centering
  \caption{Dataset statistics, including market, region, number of stocks, train/validation/test splits, annual Sharpe ratio (SR), and weekly autocorrelation (Auto Corr.).}
  \scriptsize
  \resizebox{0.48\textwidth}{!}{%
  \begin{tabular}{lccccc}
    \toprule
    \textbf{Market} & \textbf{Region} & \textbf{\# Stocks} & \textbf{Tr/Va/Te Split} & \textbf{SR} & \textbf{Auto Corr.} \\
    \midrule
    CSI300 & China & 295 & \multirow{4}{*}{\begin{tabular}[c]{@{}l@{}l@{}}Tr: 2008-2019\\Va: 2020\\Te: 2021-2024\end{tabular}} & 0.294 & 0.003 \\
    CSI500 & China & 514 & & 0.265 & 0.006  \\
    NI225 & Japan & 209 & & 0.304 & -0.012 \\
    SP500 & US & 525 & & 0.524 & -0.013 \\
    \bottomrule
  \end{tabular}
  }
  \label{tab:dataset_stats}
\end{table}

%% file: tables/main-perf.tex
\begin{table*}[ht]
    \centering
    \caption{Main results across four equity markets. Metrics: Annual Return (AR), Sharpe Ratio (SR), Calmar Ratio (CR).
    Bold: best; underline: second best. \textbf{\teacher{}} consists of seven mask-based bias-specialized teachers, while \textbf{\model{}} denotes the distilled student model.
    Results are averaged over five random seeds. Detailed efficiency analysis is provided in~\cref{subsec:efficiency}.}
    \label{tab:main_results}
    \resizebox{1.0\textwidth}{!}{%
        \begin{tabular}{lccc|ccc|ccc|ccc|ccc|c}
            \toprule
            \textbf{Markets} & \multicolumn{3}{c}{\textbf{CSI300}} & \multicolumn{3}{c}{\textbf{CSI500}} & \multicolumn{3}{c}{\textbf{NI225}} & \multicolumn{3}{c}{\textbf{SP500}} & \multicolumn{3}{c}{\textbf{Avg. across markets}} & \textbf{Efficiency} \\
            \textbf{Metrics $\rightarrow$} & AR & SR & CR & AR & SR & CR & AR & SR & CR & AR & SR & CR & AR & SR & CR & FLOPs (G) \\
           \toprule
           \multicolumn{13}{l}{\textit{\textbf{Generic Time Series SOTA}}} \\
           DLinear & 0.071 & 0.259 & 0.173 & 0.192 & 0.604 & 0.481 & 0.134 & 0.553 & 0.472 & 0.378 & 0.914 & 1.168 & 0.194 & 0.583 & 0.574 & < 0.001 \\
           TimeMixer & 0.155 & 0.642 & 0.533 & 0.170 & 0.650 & 0.412 & 0.120 & 0.538 & 0.392 & 0.203 & 0.795 & 0.732 & 0.162 & 0.656 & 0.517 & 1.123 \\
           AutoFormer & 0.155 & 0.667 & 0.592 & 0.235 & 0.921 & 0.778 & 0.124 & 0.562 & 0.449 & 0.258 & 0.873 & 0.871 & 0.193 & 0.756 & 0.673 & 2.175 \\
           PatchTST & 0.149 & 0.614 & 0.506 & 0.177 & 0.668 & 0.492 & 0.159 & 0.719 & 0.554 & 0.248 & 0.936 & 0.796 & 0.183 & 0.734 & 0.587 & 0.484 \\
           iTransformer & 0.207 & 0.801 & 0.655 & 0.307 & 1.063 & 0.972 & 0.168 & 0.737 & 0.600 & 0.297 & 0.811 & 0.967 & 0.245 & 0.853 & 0.799 & 0.611 \\
           \textbf{Ensemble} & 0.187 & 0.790 & 0.745 & 0.245 & 0.956 & 0.850 & 0.138 & 0.624 & 0.491 & 0.270 & 0.938 & 0.954 & 0.210 & 0.827 & 0.760 & 4.394 \\
           \midrule
           \multicolumn{13}{l}{\textit{\textbf{Classical Architecture}}} \\
           TCN & 0.278 & \underline{1.169} & \underline{1.231} & 0.424 & 1.465 & 1.521 & 0.148 & 0.680 & 0.576 & 0.727 & 1.482 & 2.600 & 0.394 & 1.199 & 1.482 & 0.009 \\
           GRU & 0.211 & 0.875 & 0.716 & 0.359 & 1.296 & 1.158 & 0.204 & 0.849 & 0.648 & 1.026 & 1.717 & 2.954 & 0.450 & 1.184 & 1.369 & 0.481 \\
           LSTM & 0.213 & 0.842 & 0.674 & 0.301 & 1.019 & 0.869 & 0.208 & 0.845 & 0.674 & 0.969 & \underline{1.728} & 2.612 & 0.423 & 1.109 & 1.207 & 0.639 \\
           Transformer & 0.287 & 0.915 & 0.828 & 0.622 & 1.662 & 1.747 & 0.192 & 0.774 & 0.643 & 1.254 & 1.297 & 1.881 & 0.589 & 1.162 & 1.275 & 0.810 \\
           Mamba & 0.170 & 0.648 & 0.536 & 0.254 & 0.842 & 0.645 & \underline{0.239} & \textbf{0.972} & \underline{0.781} & 1.412 & \textbf{1.834} & \textbf{3.861} & 0.519 & 1.074 & 1.456 & 0.170 \\
           \textbf{Ensemble} & 0.291 & 1.061 & 0.992 & 0.531 & 1.667 & 1.748 & 0.228 & 0.922 & 0.757 & 1.295 & 1.692 & \underline{3.193} & 0.586 & 1.336 & 1.673 & 2.109 \\
           \midrule
           \multicolumn{13}{l}{\textit{\textbf{Financial Time Series Specialist}}} \\
           RankLSTM & 0.205 & 0.753 & 0.663 & 0.272 & 0.869 & 0.699 & 0.152 & 0.610 & 0.447 & 0.753 & 1.414 & 2.138 & 0.346 & 0.912 & 0.987 & 0.639 \\
           MASTER & 0.167 & 0.675 & 0.537 & 0.203 & 0.740 & 0.458 & 0.148 & 0.637 & 0.511 & 0.660 & 1.281 & 1.882 & 0.295 & 0.833 & 0.847 & 8.725 \\
           StockMixer & 0.170 & 0.692 & 0.518 & 0.183 & 0.662 & 0.487 & 0.071 & 0.322 & 0.251 & 0.272 & 0.843 & 0.832 & 0.174 & 0.630 & 0.522 & 0.005 \\
           AlphaMix & 0.092 & 0.350 & 0.217 & 0.336 & 1.165 & 0.956 & 0.189 & 0.810 & 0.645 & 0.544 & 1.302 & 1.655 & 0.290 & 0.907 & 0.868 & 0.482 \\
           MERA & 0.352 & 1.068 & 0.889 & 0.194 & 0.478 & 0.233 & 0.135 & 0.578 & 0.421 & 0.384 & 1.041 & 1.097 & 0.266 & 0.792 & 0.660 & 9.480 \\
           \textbf{Ensemble} & 0.170 & 0.694 & 0.513 & 0.237 & 0.866 & 0.558 & 0.131 & 0.566 & 0.456 & 0.512 & 1.316 & 1.795 & 0.263 & 0.860 & 0.831 & 19.331 \\
           \midrule
           \multicolumn{13}{l}{\textit{\textbf{Ours}}} \\
           \teacher{} & \underline{0.353} & 1.152 & 1.163 & \underline{0.715} & \underline{1.883} & \underline{2.270} & 0.233 & 0.902 & 0.729 & \underline{1.511} & 1.665 & 2.791 & \underline{0.703} & \underline{1.401} & \underline{1.738} & 6.032 \\
            \textbf{\model{} (Distilled Student)} & \textbf{0.469} & \textbf{1.343} & \textbf{1.523} & \textbf{0.986} & \textbf{2.010} & \textbf{2.466} & \textbf{0.261} & \underline{0.958} & \textbf{0.784} & \textbf{1.913} & 1.506 & 2.965 & \textbf{0.907} & \textbf{1.454} & \textbf{1.934} & 0.810 \\
           \bottomrule
        \end{tabular}}
\end{table*}

%% file: tables/ablation-supervision.tex
\begin{table}[t]
    \centering
    \caption{Ablation of supervision ratio. Results report Sharpe Ratio (SR) 
    under different weightings of the distillation and ground-truth ranking 
    targets. Bold: best; underline: second best.}
    \resizebox{0.45\textwidth}{!}{%
        \begin{tabular}{lcccc|c}
            \toprule
            \textbf{Supervision} & \textbf{CSI300} & \textbf{CSI500} & \textbf{NI225} & \textbf{SP500} & \textbf{Avg. SR} \\
            \midrule
            $0.5 \times \text{distill} + 0.5 \times \text{target}$ & \textbf{1.403} & \textbf{2.080} & 0.883 & 1.462 & \underline{1.457} \\
            $0.9 \times \text{distill} + 0.1 \times \text{target}$ & 1.307 & \underline{2.036} & \textbf{0.985} & \underline{1.505} & \textbf{1.458} \\
            $1.0 \times \text{distill}$ (\model{}) & \underline{1.343} & 2.010 & \underline{0.958} & \textbf{1.506} & 1.454 \\
            \bottomrule
        \end{tabular}
    }
    \label{tab:ablation_supervision}
\end{table}

%% file: tables/ablation-distill.tex
\begin{table}[t]
    \centering
   \caption{Ablation of distillation regularization components.
Results report Sharpe Ratio (SR) obtained by incrementally adding regularization terms to vanilla distillation. SR Gap denotes the absolute train-test SR difference averaged over markets. Bold: best; underline: second best.}
    \resizebox{0.45\textwidth}{!}{%
        \begin{tabular}{lcccc|cc}
            \toprule
            \textbf{Method} & \textbf{CSI300} & \textbf{CSI500} & \textbf{NI225} & \textbf{SP500} & \textbf{Avg. SR} & \textbf{SR Gap ($\downarrow$)}\\
            \midrule
            \teacher{} & 1.152 & 1.883 & 0.902 & \textbf{1.665} & 1.401 & 3.172 \\
            \midrule
            \textit{\textbf{\model{}}} & & & & & \\
            Vanilla Distill & 1.203 & 1.826 & 0.882 & \underline{1.573} & 1.371 & \textbf{0.341} \\
            + Low Temp. & \textbf{1.440} & 1.810 & \textbf{0.973} & 1.473 & 1.424 & 0.546 \\
            + Aggressive LS & 1.324 & \textbf{2.024} & 0.905 & 1.554 & \underline{1.452} & 0.560 \\
            + SWA & \underline{1.343} & \underline{2.010} & \underline{0.958} & 1.506 & \textbf{1.454} & \underline{0.436} \\
            \bottomrule
        \end{tabular}
    }
    \label{tab:ablation_reg}
\end{table}


%% file: tables/score-concentration.tex
\begin{table}[t]
    \centering
    \caption{Average top-1 portfolio weight as a proxy for extreme-score concentration across distillation variants and individual teacher groups.}
    \resizebox{0.35\textwidth}{!}{%
        \begin{tabular}{lcccc}
            \toprule
            \textbf{Method} & \textbf{CSI300} & \textbf{CSI500} & \textbf{NI225} & \textbf{SP500} \\
            \midrule
            Causality              & 0.257 & 0.295 & 0.240 & 0.447 \\
            Locality               & 0.236 & 0.241 & 0.238 & 0.369 \\
            Periodicity            & 0.235 & 0.244 & 0.243 & 0.373 \\
            Vanilla                & 0.239 & 0.250 & 0.253 & 0.413 \\
            \midrule
            \model{} w/o LS \& SWA & 0.563 & 0.655 & 0.470 & 0.641 \\
            \model{} w/o SWA       & 0.359 & 0.414 & 0.315 & 0.458 \\
            \model{}               & 0.361 & 0.413 & 0.304 & 0.468 \\
            \bottomrule
        \end{tabular}
    }
    \label{tab:score_concentration}
\end{table}

%% file: tables/student-teacher-similarity.tex
\begin{table}[t]
\centering
\caption{Teacher--student rank similarity across distillation regularization stages. Values report mean Spearman correlation $\pm$ standard deviation computed over the seven teachers.
}
\resizebox{\linewidth}{!}{%
\begin{tabular}{lccccc}
\toprule
\textbf{Distill Steps} & \textbf{CSI300} & \textbf{CSI500} & \textbf{NI225} & \textbf{SP500} & \textbf{Average} \\
\midrule
Vanilla Distill & $0.607 \pm 0.059$ & $0.658 \pm 0.064$ & $0.342 \pm 0.282$ & $0.517 \pm 0.311$ & 0.531 $\pm$ 0.179 \\
+ Low Temp. & $0.543 \pm 0.054$ & $0.584 \pm 0.047$ & $0.252 \pm 0.190$ & $0.433 \pm 0.232$ & 0.453 $\pm$ 0.131 \\
+ Aggressive LS & $0.472 \pm 0.040$ & $0.488 \pm 0.022$ & $0.251 \pm 0.190$ & $0.283 \pm 0.104$ & 0.374 $\pm$ 0.089 \\
+ SWA & \textbf{0.454 $\pm$ 0.035} & \textbf{0.460 $\pm$ 0.037} & \textbf{0.238 $\pm$ 0.190} & \textbf{0.297 $\pm$ 0.125} & \textbf{0.362 $\pm$ 0.097} \\
\midrule
\textit{Total Reduction} & \textit{-25.2\%} & \textit{-30.1\%} & \textit{-30.6\%} & \textit{-42.6\%} & \textit{-31.8\%} \\
\bottomrule
\end{tabular}
}
\label{tab:similarity_master}
\end{table}

%% file: tables/ablation-mask.tex
\begin{table}[t]
    \centering
    \caption{Ensemble vs.\ distillation across different bias configurations. \textbf{Bold}: better method per bias setting and market.}
    \resizebox{0.48\textwidth}{!}{%
        \begin{tabular}{llcccc|c}
            \toprule
            \textbf{Bias Configuration} & \textbf{Method} & \textbf{CSI300} & \textbf{CSI500} & \textbf{NI225} & \textbf{SP500} & \textbf{Avg. SR} \\
            \midrule
            \multirow{2}{*}{Causality only} & Ensemble & \textbf{1.250} & 1.868 & 0.877 & \textbf{1.636} & 1.408 \\
             & Distill  & 1.235 & \textbf{2.082} & \textbf{1.010} & 1.476 & \textbf{1.451} \\
            \midrule
            \multirow{2}{*}{Locality only} & Ensemble & 1.118 & 1.813 & 0.754 & 1.453 & 1.285 \\
            & Distill  & \textbf{1.215} & \textbf{1.815} & \textbf{0.904} & \textbf{1.486} & \textbf{1.355} \\
            \midrule
            \multirow{2}{*}{Periodicity only} & Ensemble & 1.155 & 1.774 & 0.788 & \textbf{1.487} & 1.301 \\
           & Distill  & \textbf{1.232} & \textbf{1.828} & \textbf{0.793} & 1.449 & \textbf{1.326} \\
           \midrule
            \multirow{2}{*}{All Biases} & Ensemble & 1.152 & 1.883 & 0.902 & \textbf{1.665} & 1.401 \\
              & Distill  & \textbf{1.343} & \textbf{2.010} & \textbf{0.958} & 1.506 & \textbf{1.454} \\
            \bottomrule
        \end{tabular}
    }
    \label{tab:ablation_bias}
\end{table}

%% file: content/discussion.tex
\section{Understanding the Synthesis Mechanism}
\label{sec:analysis}

Although \model{} achieves state-of-the-art performance (Table~\ref{tab:main_results}), the aggressive regularization employed during distillation substantially reduces teacher--student similarity (\cref{tab:similarity_master}).
This discrepancy indicates that the distilled student does not simply replicate teacher predictions. To characterize the resulting synthesis behavior, we conduct three complementary analyses.
First, in \cref{sec:statistical_validation}, we examine whether \model{} produces statistically significant excess returns beyond those of its teachers.
Second, in \cref{sec:attention_alignment}, we assess whether the student collapses to a static average of its teachers or dynamically adapts its attention behavior across days.
Third, in \cref{sec:behavioral_adaptation}, we analyze whether \model{} exhibits regime-adaptive behavior by aligning more closely with different inductive biases under different market conditions, and connect this behavior to empirical performance.
Across all analyses, we adopt a consistent evaluation protocol: for each market, we train five models with different random initializations and conduct statistical tests on aggregated predictions, with reported $p$-values computed on the averaged series.

\subsection{Statistical Validation of Excess Returns}
\label{sec:statistical_validation}

We quantify the synthesis effect through a performance attribution analysis by regressing \model{} returns on baseline returns.
Table~\ref{tab:alpha_analysis} reports the estimated daily Alpha ($\alpha$), $t$-statistics of Alpha, Beta ($\beta$), and $R^2$ values.
When compared with the vanilla Transformer, \model{} achieves statistically significant positive Alpha across all markets
($\alpha = 0.26\sim1.03 \times 10^{-3}$), reflecting returns not explained by unconstrained bidirectional attention. This suggests that incorporating inductive biases through distillation improves signal extraction in non-stationary markets.

More importantly, \model{} also yields statistically significant Alpha when benchmarked against its own teacher (\teacher{}) on CSI300 and CSI500
($\alpha = 0.50\sim0.63 \times 10^{-3}$).
This student-surpasses-teacher effect indicates that distillation goes beyond direct teacher imitation, instead integrating information across bias-specialized teachers.
Meanwhile, the estimated Beta coefficients remain close to unity ($\beta = 0.97\sim1.32$), with moderate $R^2$ values ($0.65\sim0.77$),
suggesting that \model{} preserves overall directional exposure while refining relative ranking decisions.

This raises a natural question: does the student's behavioral similarity to its teachers reflect genuine synthesis, or merely collapse to a static average of teacher predictions?

\input{tables/prove-excess-return}

\input{tables/attention-behavior}

\subsection{Attention-Level Teacher Alignment}
\label{sec:attention_alignment}

To assess this, we evaluate the student at the level of daily cross-sectional stock selection: for each test day, we aggregate the two-layer attentions of the top-5 selected stocks using portfolio weights, then assign the resulting attention pattern to the most similar teacher group or an explicit average-teacher baseline via cosine similarity.

Table~\ref{tab:attention_behavior} shows that the student is rarely closest to the average teacher, while its nearest teacher group varies across days and markets. This suggests that \model{} does not default to a fixed average of teacher behaviors, motivating a finer-grained analysis of how the student aligns with different inductive biases under varying market conditions, which we investigate in \cref{sec:behavioral_adaptation}.

\input{tables/input_level_analysis}

\subsection{Evidence of Conditional Bias Activation}
\label{sec:behavioral_adaptation}

We investigate whether \model{} exhibits regime-dependent behavior consistent with different inductive biases, which we refer to as \emph{conditional bias activation}, by measuring its portfolio weighted similarity to classical baseline architectures (e.g., GRU, TCN, and Mamba).
Although \model{} is not explicitly trained to imitate these baselines, its bias-specialized teachers implicitly encode inductive priors commonly associated with such architectures.
Similarity analysis therefore serves as a diagnostic tool for assessing whether the distilled student selectively aligns with distinct temporal behaviors under different market conditions.

\paragraph{\textbf{Defining Daily Strategy Similarity.}}
To analyze conditional bias activation, we summarize each model’s daily trading behavior using a portfolio weighted temporal representation constructed from the 5-day moving average feature, which reflects the effective \emph{input pattern} emphasized by the model at each trading day.
Daily strategy similarity between \model{} and each baseline is then measured via cosine similarity between their respective representations.
Full definitions and implementation details are provided in \cref{subsec:pattern_similarity}.

To assess conditional behavior, we partition trading days into \textit{good} (top 30\%) and \textit{bad} (bottom 30\%) regimes and measure the regime-wise difference in similarity, denoted as $\Delta\rho$.
A significant $\Delta\rho$ indicates regime-dependent alignment with a given inductive bias.

\paragraph{\textbf{Conditional and Market-Dependent Bias Activation.}}
As shown in Table~\ref{tab:conditional_pattern},
\model{} exhibits statistically significant conditional alignment with certain inductive biases over time, rather than uniformly imitating any single baseline.
For example, \model{} shows consistent positive conditional similarity with GRU across markets ($p < 0.05$), indicating increased alignment with sequential dependencies during periods when such patterns are beneficial, even when GRU is not globally competitive.
At the same time, the strength and consistency of conditional alignment vary across markets:
in SP500 and NI225, \model{} displays broader and more stable alignment with sequential models and TCN, whereas such patterns are weaker or less consistent in CSI markets.
This cross-market heterogeneity suggests that \model{}’s bias-aligned behavior depends on market context rather than a fixed preference for any single inductive bias.

Taken together, these analyses reconcile the reduced teacher--student similarity observed during training (\cref{tab:similarity_master}) with the strong empirical performance of \model{} (\cref{tab:main_results}).
Rather than reproducing teacher predictions, \model{} exhibits behavior indicative of when different inductive biases are beneficial, yielding adaptive behavior that goes beyond static ensembling.

%% file: tables/prove-excess-return.tex
\begin{table}[t]
\centering
\caption{Performance attribution analysis. We regress \model{} returns on vanilla Transformer and \teacher{} returns, reporting daily Alpha ($\alpha$), $t$-statistics of Alpha, Beta ($\beta$), and $R^2$. $^{*}p < 0.05$, $^{**}p < 0.01$, $^{***}p < 0.001$ (one-tailed).}
\label{tab:alpha_analysis}
\resizebox{\linewidth}{!}{%
\begin{tabular}{l|clcc|clcc}
\toprule
 & \multicolumn{4}{c|}{\textbf{vs. Vanilla Transformer}} & \multicolumn{4}{c}{\textbf{vs. \teacher{}}} \\
\textbf{Market} & \textbf{Alpha ($\alpha$)} & \textbf{$t$-stat} & \textbf{Beta ($\beta$)} & \textbf{$R^2$} & \textbf{Alpha ($\alpha$)} & \textbf{$t$-stat} & \textbf{Beta ($\beta$)} & \textbf{$R^2$} \\
\midrule
CSI300  & $0.75 \times 10^{-3}$ & $3.724^{***}$ & 0.975 & 0.585 & $0.50 \times 10^{-3}$ & $2.706^{**}$ & 0.973 & 0.651 \\
CSI500  & $0.93 \times 10^{-3}$ & $3.711^{***}$ & 1.207 & 0.682 & $0.63 \times 10^{-3}$ & $2.780^{**}$ & 1.156 & 0.742 \\
NI225   & $0.26 \times 10^{-3}$ & $1.965^{*}$   & 1.019 & 0.703 & $0.12 \times 10^{-3}$ & 1.056        & 0.988 & 0.768 \\
SP500   & $1.03 \times 10^{-3}$ & $1.655^{*}$   & 1.319 & 0.698 & $0.21 \times 10^{-3}$ & 0.341        & 1.231 & 0.705 \\
\bottomrule
\end{tabular}
}
\end{table}

%% file: tables/attention-behavior.tex
\begin{table}[t]
    \centering
    \caption{Fraction of test days on which the student attention map is closest 
    to each teacher group or the average-teacher baseline, measured via cosine 
    similarity over portfolio-weighted two-layer attentions of the top-5 selected 
    stocks.
    }
    \resizebox{0.3\textwidth}{!}{%
        \begin{tabular}{lcccc}
            \toprule
            \textbf{Teacher Group} & \textbf{CSI300} & \textbf{CSI500} & \textbf{NI225} & \textbf{SP500} \\
            \midrule
            Causality   & 0.090 & 0.258 & 0.000 & 0.001 \\
            Locality    & 0.319 & 0.280 & 0.323 & 0.313 \\
            Periodicity & 0.409 & 0.337 & 0.373 & 0.434 \\
            Vanilla     & 0.166 & 0.104 & 0.238 & 0.147 \\
            \midrule
            Average Teacher & 0.017 & 0.022 & 0.066 & 0.105 \\
            \bottomrule
        \end{tabular}
    }
    \label{tab:attention_behavior}
\end{table}

%% file: tables/input_level_analysis.tex
\begin{table}[t]
\centering
\caption{Conditional alignment analysis using MA5 features.
$\bar{\rho}$ denotes overall pattern similarity, and $\Delta\rho$ measures the increase in similarity during profitable periods (top 30\% returns) relative to unprofitable periods.
\textbf{Bold}/\underline{underline} indicate the best/second-best Sharpe ratio (SR) per market. $^{*}p < 0.05$, $^{**}p < 0.01$, $^{***}p < 0.001$ (one-tailed).}
\label{tab:conditional_pattern}
\scriptsize
\resizebox{\linewidth}{!}{%
\begin{tabular}{llccccc}
\toprule
\textbf{Market} & \textbf{Baseline} & \textbf{SR} & \textbf{$\bar{\rho}$} & \textbf{$\Delta\rho$} & \textbf{$t$-stat} & \textbf{$p$-value} \\
\midrule
\multirow{5}{*}{CSI300}
  & Vanilla Transformer & \underline{0.915} & 0.660 & 0.038 & 2.57 & 0.005** \\
  & LSTM       & 0.842 & 0.649 & 0.025 & 1.18 & 0.120 \\
  & GRU        & 0.875 & 0.650 & 0.049 & 2.19 & 0.014* \\
  & Mamba      & 0.648 & 0.608 & 0.055 & 2.12 & 0.017* \\
  & TCN        & \textbf{1.169} & 0.599 & 0.023 & 1.47 & 0.071 \\
\cmidrule{1-7}
\multirow{5}{*}{CSI500}
  & Vanilla Transformer & \textbf{1.662} & 0.738 & -0.012 & -0.80 & 0.787 \\
  & LSTM       & 1.019 & 0.618 & -0.004 & -0.18 & 0.570 \\
  & GRU        & 1.296 & 0.647 & 0.053 & 2.54 & 0.006** \\
  & Mamba      & 0.842 & 0.582 & -0.042 & -1.66 & 0.951 \\
  & TCN        & \underline{1.465} & 0.649 & -0.011 & -0.61 & 0.728 \\
\cmidrule{1-7}
\multirow{5}{*}{NI225}
  & Vanilla Transformer & 0.774 & 0.557 & 0.036 & 1.83 & 0.034* \\
  & LSTM       & 0.845 & 0.562 & 0.066 & 3.35 & <0.001*** \\
  & GRU        & \underline{0.849} & 0.536 & 0.046 & 2.10 & 0.018* \\
  & Mamba      & \textbf{0.972} & 0.563 & 0.075 & 3.61 & <0.001*** \\
  & TCN        & 0.680 & 0.483 & 0.041 & 1.96 & 0.025* \\
\cmidrule{1-7}
\multirow{5}{*}{SP500} 
  & Vanilla Transformer & 1.297 & 0.613 & -0.004 & -0.20 & 0.578 \\
  & LSTM       & \underline{1.728} & 0.567 & 0.065 & 2.51 & 0.006** \\
  & GRU        & 1.717 & 0.574 & 0.087 & 3.28 & <0.001*** \\
  & Mamba      & \textbf{1.834} & 0.559 & 0.051 & 1.94 & 0.027* \\ 
  & TCN        & 1.482 & 0.427 & 0.058 & 2.10 & 0.018* \\
  \bottomrule
\end{tabular}%
}
\end{table}

%% file: content/related.tex
\section{Related Works}

\paragraph{\textbf{Transformers for Generic and Financial Time Series Forecasting}}
Transformers~\citep{vaswani2017attention} have been widely adopted for time series
forecasting due to their ability to capture long-range dependencies beyond
RNNs~\citep{hochreiter1997long}. Early variants addressed the quadratic cost of
self-attention via sparse designs, including LogTrans~\citep{li2019enhancing} and
Informer~\citep{zhou2021informer}. Subsequent models incorporated inductive biases
tailored to generic time series, such as decomposition- and frequency-based designs
(Autoformer, FEDformer)~\citep{wu2021autoformer,yi2023frequency} and locality- or
inter-variable-aware architectures (PatchTST, iTransformer)~\citep{nie2022time,liu2023itransformer}. In financial forecasting, Transformers have been applied to jointly model temporal dynamics and cross-sectional stock relationships~\citep{yoo2021accurate,li2024master}.
To address market heterogeneity, prior work explored adaptive designs including routing mechanisms~\citep{lin2021learning}, graph-based models~\citep{hsu2021fingat,xia2024ci}, and Mixture-of-Experts frameworks~\citep{sun2023mastering,yu2024miga,liu2025mera}. While these approaches introduce adaptivity through input conditioning or expert routing, the inductive biases governing temporal dependency modeling are typically fixed during training.

\paragraph{\textbf{Inductive Biases of Transformers}}
Although originally designed with minimal structural assumptions~\citep{vaswani2017attention}, Transformers have been shown to exhibit implicit inductive preferences~\citep{tay2023scaling}. In sequential domains, modeling often requires directional, local, and structured temporal priors, whereas standard Transformers favor global content-based interactions unless such biases are explicitly encoded~\citep{tay2021pre}. This has motivated structured attention mechanisms and positional designs, including relative positional bias (RPB)~\citep{raffel2020exploring}, ALiBi~\citep{press2021train}, and periodic-aware attention~\citep{sun2025penguin}. More recent work has incorporated inductive biases inspired by recurrent state evolution and convolutional processing~\citep{huang2022encoding,katharopoulos2020transformers,dosovitskiy2020image,nie2022time}, suggesting a unified view of sequence modeling through shared inductive priors.

\paragraph{\textbf{Knowledge Distillation with Transformers}}
Knowledge Distillation (KD)~\citep{hinton2015distilling} is a standard paradigm for
compressing Transformer models.
Early approaches such as DeiT~\citep{touvron2021training} introduced task-specific
distillation tokens, while subsequent methods, including TinyBERT~\citep{jiao2020tinybert}
and Squeezing-Heads Distillation~\citep{bing2025optimizing}, address architectural
mismatch via attention-level alignment.
In time series settings, TimeKD~\citep{liu2025efficient} distills privileged
representations from language models for multivariate forecasting, and AnomalyLLM~\citep{liu2024large}
adapts KD to anomaly detection by transferring knowledge from pretrained LLMs while
preserving teacher–student discrepancies.

%% file: content/conclude.tex
\section{Conclusion}

This work investigates the role of inductive biases in Transformer-based financial time series modeling and demonstrates that diverse temporal priors—including causality, locality, and periodicity—can be effectively integrated within a single model through knowledge distillation.
We propose \textbf{TIPS}, a framework that trains bias-specialized Transformer teachers via attention masking and distills their collective knowledge into a compact student model that exhibits regime-dependent utilization of different inductive biases.
Extensive experiments across multiple equity markets, supported by fine-grained ablation and behavioral analyses, show that TIPS achieves state-of-the-art performance while substantially reducing inference-time computational cost relative to ensemble baselines.
Beyond empirical gains, our analyses reveal that TIPS does not rely on rigid architectural mimicry; instead, it exhibits behavior aligned with different inductive biases under varying market conditions, enabling robust generalization under pronounced non-stationarity.
These findings highlight conditional bias utilization as a principled perspective for building adaptive and efficient forecasting systems in dynamic environments.

Future work includes extending TIPS to intermediate representation distillation, parameter-efficient teacher sharing, and explicit regime-aware bias adaptation that builds on the emergent conditional bias utilization observed in this work, as well as applying the framework to other non-stationary domains such as energy, traffic, and climate forecasting.

%% file: content/appendix.tex
\clearpage
\section{Experimental Details}

\subsection{Definition of Input Features and Label}
\label{subsec:features}

We generate 8 temporal features: 5 rolling z-score normalized OHLCV features and 3 moving average features. Each OHLCV feature is normalized independently using the same 20-day rolling z-score (exemplified by close price below). 
For moving averages $z_{d_k}^t$ with $k \in \{5, 10, 20\}$, we compute the ratio of the $k$-day moving average to the current close price, minus one.
The label $y_t$ is the $q$-day cumulative return with $q = 5$:

\begin{equation*}
\resizebox{\linewidth}{!}{$
z_{\text{close}}^t = \dfrac{x_{\text{close}}^t - \bar{x}_{\text{close}}^{[t-19:t]}}{\sigma_{x_{\text{close}}}^{[t-19:t]}}, \quad
z_{d_k}^t = \dfrac{\frac{1}{k}\sum_{i=0}^{k-1} x_{\text{close}}^{t-i}}{x_{\text{close}}^t} - 1, \quad
y_t = \dfrac{x_{\text{close}}^{t+q-1} - x_{\text{close}}^t}{x_{\text{close}}^t}
$}
\end{equation*}

\subsection{Hyperparameter Configurations}
\label{subsec:baseline}

\paragraph{\textbf{Baselines.}}
For classical architectures (GRU, LSTM, Mamba, TCN, Transformer), 
TimeMixer, and PatchTST, we follow FinTSB~\citep{hu2025fintsb}.
For StockMixer, MASTER, AlphaMix, and MERA, we use configurations 
from their original implementations. All other models use default 
settings from the Time-Series Library. RankLSTM uses identical 
hyperparameters to our LSTM baseline. Hyperparameter configurations 
are summarized in Table~\ref{tab:hyperparam}.

\begin{table}[H]
\centering
\caption{Baseline hyperparameter configurations.}
\resizebox{0.45\textwidth}{!}{%
\begin{tabular}{lccccc}
\toprule
\textbf{Model} & \textbf{Hidden} & \textbf{Layers} & \textbf{FFN} & \textbf{Heads} & \textbf{Dropout} \\
\midrule
GRU / LSTM          & 64 & 2 & —   & — & 0.0 \\
TCN$^\dagger$        & 64 & 2 & —   & — & 0.0 \\
Transformer         & 64 & 2 & 256 & 4 & 0.0 \\
PatchTST$^\ddagger$ & 64 & 2 & 256 & — & 0.0 \\
iTransformer        & 64 & 2 & 256 & — & 0.0 \\
AutoFormer          & 64 & 2 & —   & — & 0.1 \\
TimeMixer$^\P$      & 64 & 2 & 256 & — & 0.0 \\
Mamba$^\S$          & 32 & 1 & 64  & — & 0.1 \\
\bottomrule
\end{tabular}}
\begin{minipage}{0.45\textwidth}
\footnotesize
$^\dagger$ TCN: kernel size 4. \\
$^\ddagger$ PatchTST: patch length 2, stride 1, factor 3. \\
$^\S$ Mamba: state size 16, conv size 4, expand factor 2. \\
$^\P$ TimeMixer: down-sampling window 2, moving-average 25, 3 down-sampling layers, \texttt{moving\_avg} decomposition, channel independence \texttt{True}.
\end{minipage}
\label{tab:hyperparam}
\end{table}

\paragraph{\textbf{TIPS Teachers}}
Most bias-specialized teachers ($\mathcal{T}_{\text{ALiBi}}$, $\mathcal{T}_{\text{fixed}}$, $\mathcal{T}_{\text{learn}}$, etc.) introduce inductive biases at the attention level. For the ALiBi mask, we assign decay rates $\{2^{-8}, 2^{-4}, 2^{-8/3}, 2^{-2}\}$ to the four attention heads respectively. For the PENGUIN mask, we assign periods $p \in \{5, 10, 15, 20\}$ to the four attention heads respectively, aligning with weekly, biweekly, and monthly cycles within the 20-day lookback window. In contrast, the patch-based teacher ($\mathcal{T}_{\text{patch}}$) introduces inductive bias at input level through temporal segmentation, requiring two additional hyperparameters to generate overlapping patches by setting patch length to $2$ and stride to $1$.  All other Transformer architecture hyperparameters follow the original settings detailed above.

\paragraph{\textbf{TIPS Students}}
The student is trained for 20 epochs using the Adam optimizer with learning rate $\eta = 10^{-4}$. We apply low temperature distillation with $\tau = 0.01$ to preserve fine-grained ranking signals, and aggressive label smoothing with $\varepsilon = 0.9$ to decouple ranking structure from prediction calibration and mitigate overfitting to individual teacher behaviors. Stochastic weight averaging (SWA) is applied over the final 10 epochs to bias optimization toward flatter minima and improve robustness under regime shifts.

\subsection{Training Details}
\label{subsec:implement-detail}

We train all models using the Adam optimizer~\citep{kingma2014adam} with learning rate $\eta=10^{-3}$ for classical architectures and $\eta=10^{-4}$ for generic time series SOTA models and financial time series specialists, with an effective batch size of $256$ via gradient accumulation. 
All baseline models and bias-specialized teachers are trained for $100$ epochs, and the student model for $20$ epochs. Models are implemented in PyTorch and trained on NVIDIA RTX 3090 GPUs. We report the average performance over $5$ random seeds $\{0, 1, 2, 3, 4\}$.

\subsection{Evaluation Methods}
\label{subsec:eval-method}

We compute portfolio returns using a sliding window approach with top-$k$ stock selection and softmax-weighted portfolio construction. 
Given daily predictions $\bm{P} \in \mathbb{R}^{B \times S}$ and next-day returns $\bm{R} \in \mathbb{R}^{B \times S}$, for each day $d$ we select the top-$k$ stocks by predicted value, deduplicating ties via unique selection, compute softmax weights, and accumulate weighted returns over a window of $\min(W, B-d)$ days via a circular buffer. 
The output $\bm{r} \in \mathbb{R}^{W \times B}$ is averaged across all starting days $w \in W$ to obtain the metrics. 
Transaction costs are excluded for consistency with existing benchmarks; see~\cref{sec:tcost} for their effect.

\begin{algorithm}[H]
\caption{Portfolio Returns Calculation with Sliding Window}
\label{alg:portfolio_returns}
\begin{algorithmic}[1]
\Require Predictions $\bm{P} \in \mathbb{R}^{B \times S}$, next-day returns $\bm{R} \in \mathbb{R}^{B \times S}$, top-$k$ stocks $k$, window length $W$
\Ensure Portfolio returns $\bm{r} \in \mathbb{R}^{W \times B}$
\State Initialize returns matrix $\bm{r} \gets \bm{0}^{W \times B}$
\For{$d = 0$ to $B-1$} \Comment{For each day in batch}
    \State $\text{row\_idx} \gets d \bmod W$
    \State $\text{time\_range} \gets \min(W, B - d)$
    \State Extract predictions $\bm{p}_d \gets \bm{P}[d, :]$ \Comment{Shape: $S$}
    \State Sort stocks: $\bm{I} \gets \text{argsort}(\bm{p}_d, \text{descending})$
    \State $\mathcal{S}_{\text{top}} \gets \text{unique}(\bm{I}[:k])$ \Comment{Top-$k$ stock indices}
    \State $\bm{p}_{\text{top}} \gets \bm{p}_d[\mathcal{S}_{\text{top}}]$ \Comment{Top-$k$ predictions}
    \State $\bm{w} \gets \text{softmax}(\bm{p}_{\text{top}})$ \Comment{Compute weights}
    \State $\bm{R}_{\text{range}} \gets \bm{R}[d:d+\text{time\_range}, \mathcal{S}_{\text{top}}]$ \Comment{Future returns}
    \State $\bm{r}_w \gets (\bm{R}_{\text{range}} \cdot \bm{w})^\top$ \Comment{Weighted returns}
    \State $\bm{r}[\text{row\_idx}, d:d+\text{time\_range}] \gets \sum_{s \in \mathcal{S}_{\text{top}}} \bm{r}_w[s]$ \Comment{Sum over stocks}
\EndFor
\State \Return $\bm{r}$
\end{algorithmic}
\end{algorithm}

\section{Performance Degrade with Transaction Costs}
\label{sec:tcost}

We model transaction costs covering brokerage commissions, exchange fees, and regulatory charges: CSI300 \& CSI500 incurs $\sim$0.006\% (buy) and $\sim$0.056\% (sell), NI225 $\sim$0.002\% on both sides, and the SP500 negligible costs ($\sim$0\%). With 5-day rebalancing, annualized costs amount to 3.12\%, 0.20\%, and 0\% respectively. 
After subtracting these costs, \model{} continues to outperform both the strongest external ensemble baseline (Classical Architecture Ensemble; Avg.\ AR/SR/CR: 0.891/\allowbreak1.385/\allowbreak1.854 vs.\ 0.570/\allowbreak1.273/\allowbreak1.611) 
and its own teacher (\teacher{}; Avg.\ AR/SR/CR: 0.891/\allowbreak1.385/\allowbreak1.854 vs.\ 0.687/\allowbreak1.338/\allowbreak1.669), confirming that the performance advantage is robust to realistic trading frictions.

\section{Efficiency Analysis}
\label{subsec:efficiency}

We present computational efficiency measurements on a single NVIDIA RTX 3090 GPU with batch size of 1 for fair comparison. DLinear is the most efficient (<0.001G FLOPs, 0.28ms), while Mamba offers a strong efficiency-capacity trade-off (0.170G FLOPs, 0.65ms). Attention-based models are costlier, with AutoFormer at 20.00ms and MERA at 9.480G FLOPs with over 1M parameters. Notably, \model{} matches the base Transformer's efficiency (0.810G FLOPs, 1.31ms) while retaining ensemble knowledge, unlike \teacher{} which incurs significant overhead (6.032G FLOPs, 9.6ms).

\input{tables/efficiency}

\section{Larger Vanilla Transformers Do Not Close the Performance Gap}
\label{subsec:scale_analysis}
To verify that \model{}'s gains cannot be attributed to model capacity along, we train larger vanilla Transformers with increased depth ($L$) and hidden dimension ($d$) and compare their SR against \model{}. As shown in \cref{tab:scale_analysis}, all larger Transformers substantially underperform \model{}, and increasing depth or width does not yield consistent improvement, confirming that the gains stem from inductive biases transferred through distillation rather than model scale.

\input{tables/scale-analysis}

\section{Formal Definition of Daily Strategy Representation in \cref{sec:behavioral_adaptation}}
\label{subsec:pattern_similarity}
For a model $m$ at trading day $d$, given input features
$\bm{X} \in \mathbb{R}^{S \times T \times F}$,
we define a daily strategy representation based on the 5-day moving average (MA5) feature as
$
\bm{z}_d^{m}
\;=\;
\sum_{s \in \mathcal{S}_d^{m}}
w_{d,s}^{m} \, \bm{X}_{s,:,f_{\mathrm{MA5}}},
$

where $\mathcal{S}_d^{m}$ denotes the set of top-$k$ stocks selected by model $m$ on day $d$, $w_{d,s}^{m}$ is the corresponding portfolio weight (with $\sum_{s \in \mathcal{S}_d^{m}} w_{d,s}^{m} = 1$), and $f_{\mathrm{MA5}}$ indexes the MA5 feature channel. Thus $\bm{X}_{s,:,f_{\mathrm{MA5}}} \in \mathbb{R}^{T}$ represents the MA5 sequence of stock $s$ over the $T$-day lookback window.
This representation summarizes the model’s effective trading behavior in the feature space and is consistent with the portfolio-based evaluation in \cref{sec:main_results}, as MA5 is included among the input features for all models.
Finally, daily similarity between models is computed using cosine similarity between their respective $\bm{z}_d^{m}$ representations.

\section{Market Regime Analysis}
\label{subsec:market_regime}

\begin{figure}[H]
  \centering
  \resizebox{0.78\columnwidth}{!}{%
    \begin{tabular}{c}
      \includegraphics[width=\columnwidth]{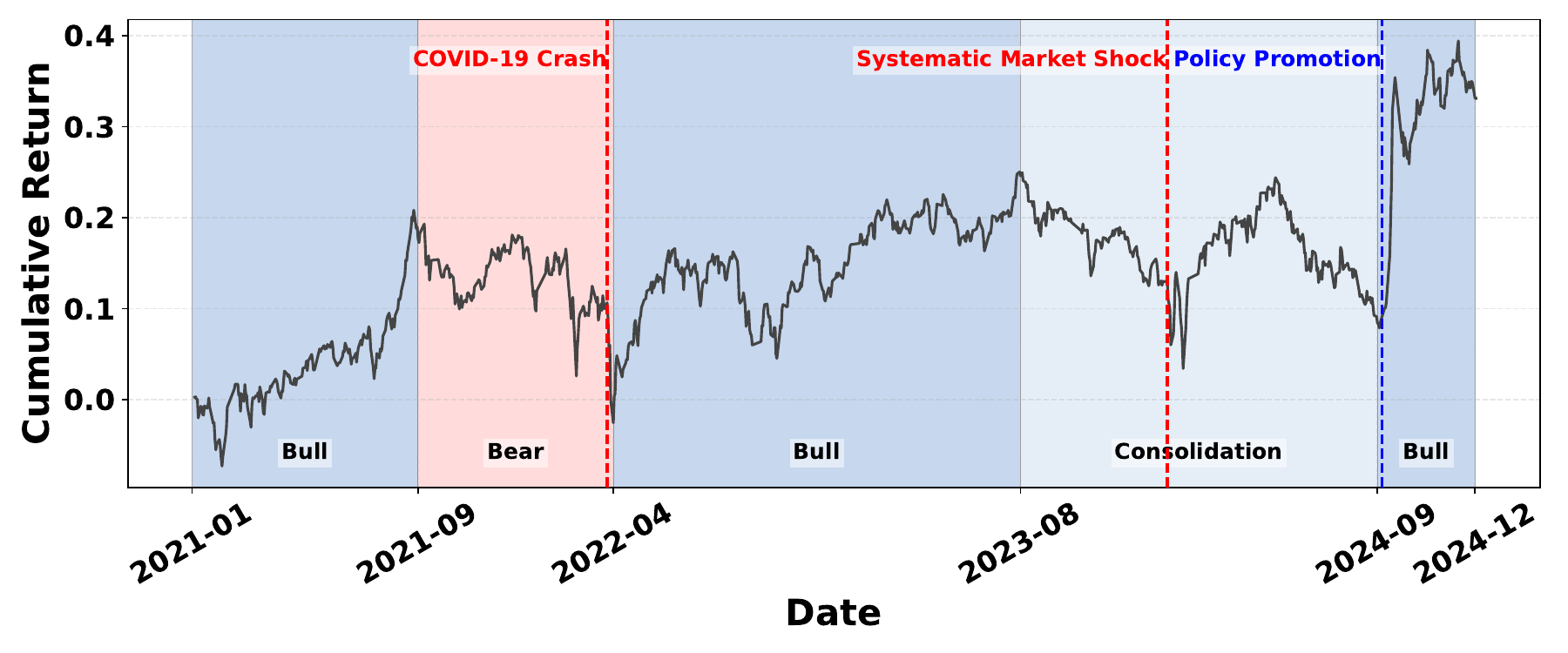}\\[0.4pt]
      \includegraphics[width=\columnwidth]{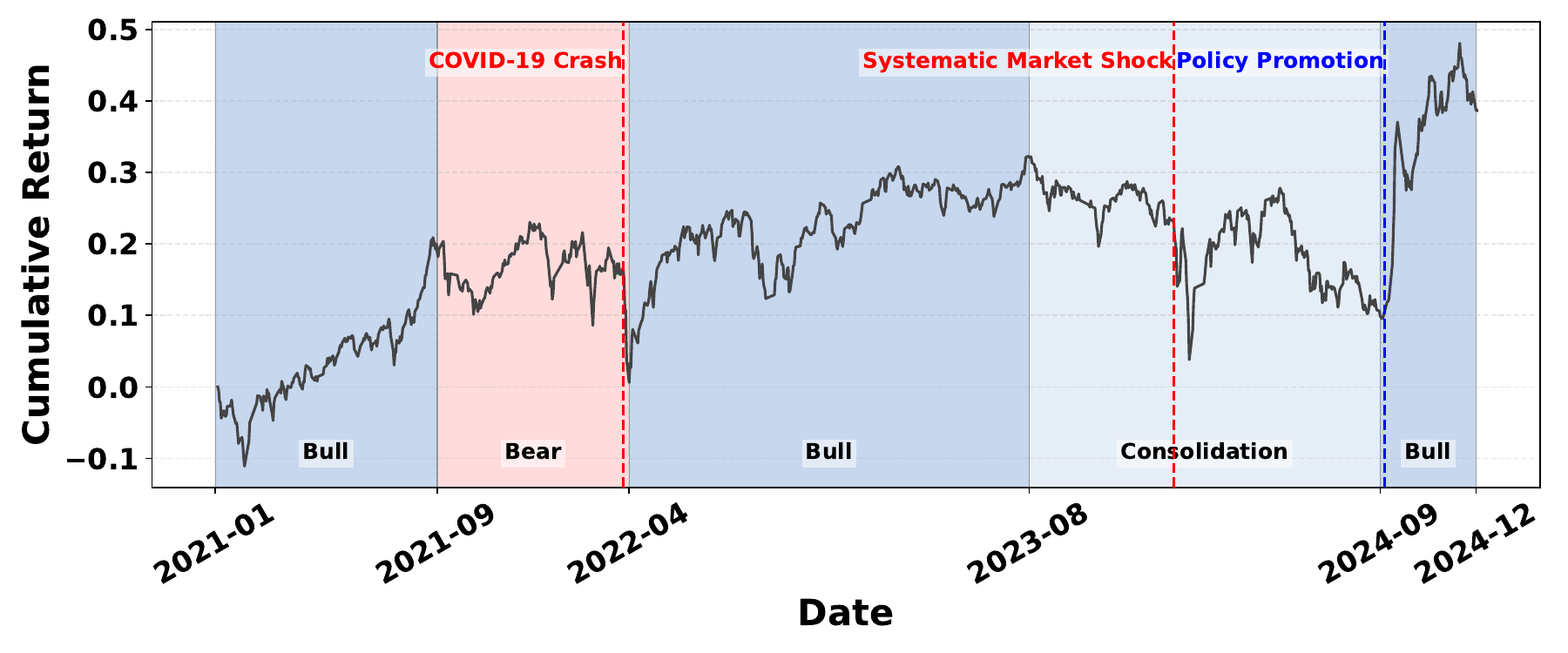}\\[0.4pt]
      \includegraphics[width=\columnwidth]{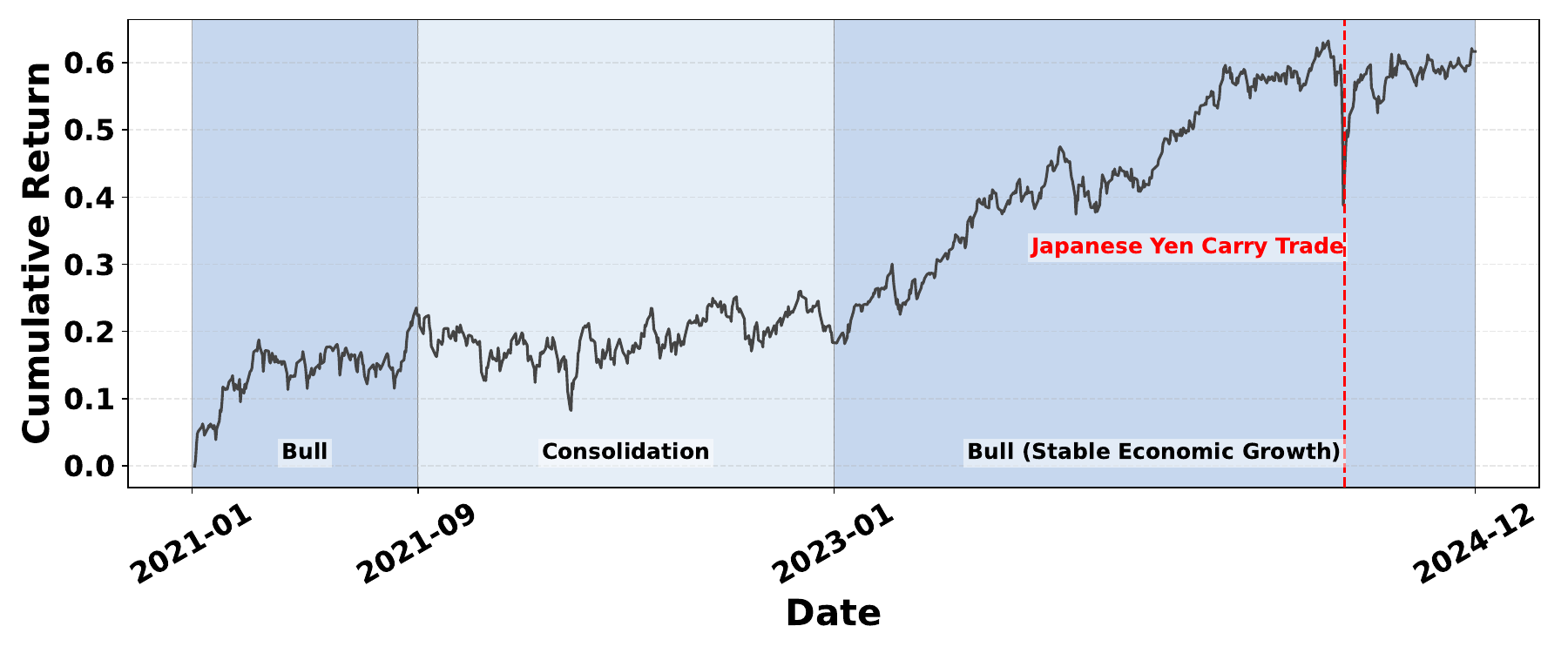}\\[0.4pt]
      \includegraphics[width=\columnwidth]{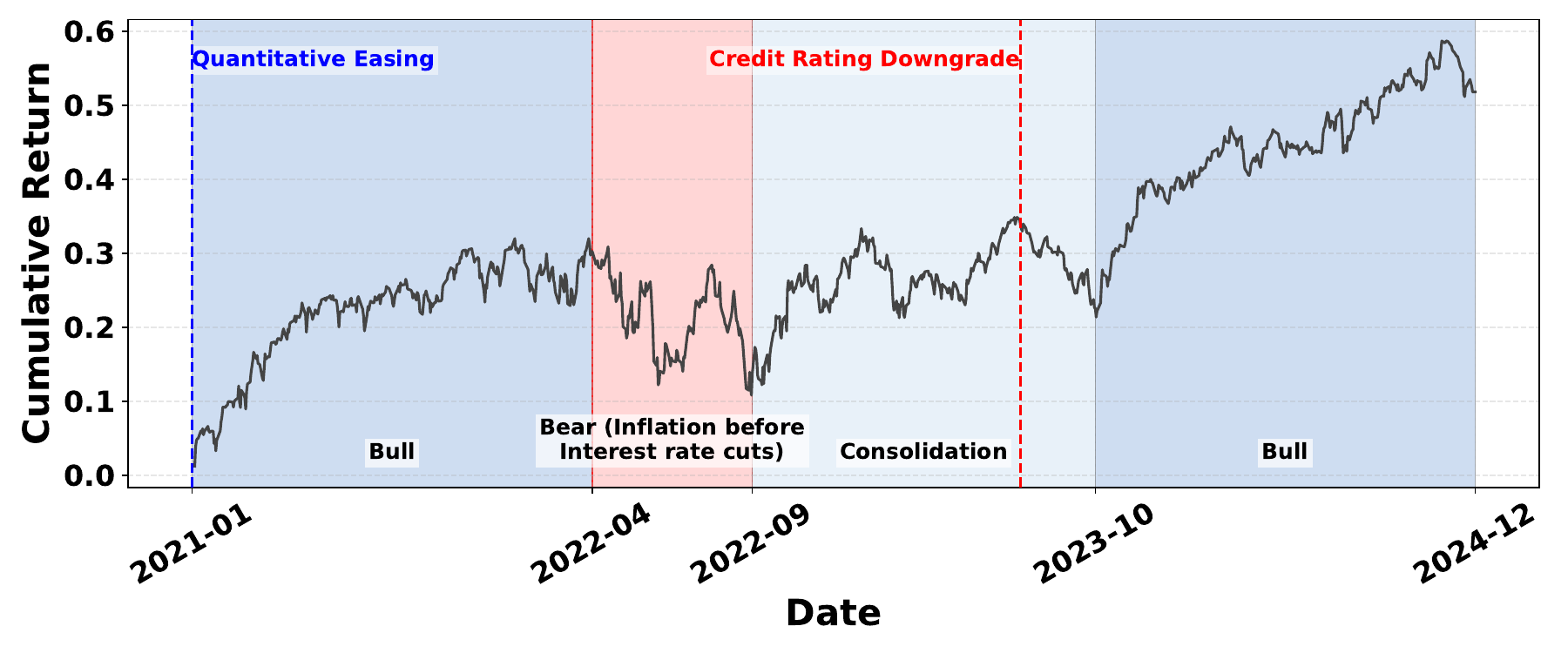}
    \end{tabular}
  }
  \caption{Regime analysis of four different markets.}
  \label{fig:regime_analysis}
\end{figure}

\Cref{fig:regime_analysis} shows market regime segmentation and cumulative returns across four equity markets (from top to bottom: CSI300, CSI500, NI225, and SP500). Regimes (\emph{Bull, Bear, Consolidation}) are aligned along the timeline, with key macro and market events annotated to aid interpretation of regime transitions.

%% file: tables/efficiency.tex
\begin{table}[H]
    \caption{Computational efficiency comparison across models. We report FLOPs (G), inference time per sample (ms), peak memory usage (MB), and total number of trainable parameters. 
    }
    \centering
    \resizebox{\columnwidth}{!}{%
    \begin{tabular}{l|cccc}
        \toprule
        \textbf{Metrics $\rightarrow$} & \textbf{FLOPs(G)} & \textbf{Inf. Time(ms)} & \textbf{Memory(MB)} & \textbf{Params} \\
        \midrule
        \multicolumn{5}{l}{\textit{\textbf{Baselines}}} \\
        DLinear & < 0.001 & 0.28 & 8.88 & 72 \\
        TimeMixer & 1.123 & 4.79 & 77.39 & 7183 \\
        AutoFormer & 2.175 & 20.00 & 46.08 & 244417 \\
        PatchTST & 0.485 & 1.43 & 15.73 & 25878 \\
        iTransformer & 0.611 & 1.07 & 28.41 & 101505\\
        GRU & 0.480 & 0.26 & 36.60 & 39233\\
        LSTM & 0.639 & 0.27 & 45.05 & 52289 \\
        Transformer & 0.810 & 1.31 & 30.67 & 104897 \\
        Mamba & 0.170 & 0.65 & 35.75 & 33409 \\
        TCN & 0.009 & 0.67 & 19.21 & 55937 \\
        RankLSTM & 0.639 & 0.27 & 45.05 & 52289 \\
        MASTER & 8.725 & 1.59 & 72.21 & 726601 \\
        StockMixer & 0.005 & 3.48 & 13.01 & 14710 \\
        AlphaMix & 0.482 & 0.21 & 36.28 & 41257 \\
        MERA & 9.480 & 8.03 & 54.73 & 1460017 \\
        Transformer RPB & 1.204 & 1.99 & 35.25 & 105025 \\
        PatchTransformer & 0.776 & 1.06 & 28.65 & 155393 \\ 
        \midrule
        \multicolumn{5}{l}{\textit{\textbf{Ours}}} \\
        \teacher{} & 6.032 & 9.6 & 217.26 & 734407 \\
        \model{} & 0.810 & 1.31 & 30.67 & 104897 \\
        \bottomrule
    \end{tabular}
    }
    \label{tab:efficiency}
\end{table}

%% file: tables/scale-analysis.tex
\begin{table}[H]
    \centering
    \caption{SR of larger vanilla Transformers compared to \model{}.}
    \resizebox{0.36\textwidth}{!}{%
        \begin{tabular}{lcccccc}
            \toprule
            & \multicolumn{4}{c}{\textbf{Larger Transformer $(L, d)$}} & \\
            \cmidrule(lr){2-5}
            \textbf{Market} & $(3,128)$ & $(3,256)$ & $(4,128)$ & $(4,256)$ & \model{} \\
            \midrule
            CSI300 & 0.65 & 0.52 & 0.45 & 0.61 & \textbf{1.34} \\
            CSI500 & 1.02 & 0.57 & 0.58 & 0.50 & \textbf{2.01} \\
            NI225  & 0.76 & 0.68 & 0.70 & 0.70 & \textbf{0.96} \\
            SP500  & 1.07 & 1.13 & 1.11 & 1.20 & \textbf{1.61} \\
            \bottomrule
        \end{tabular}
    }
    \label{tab:scale_analysis}
\end{table}

%% file: main.bib
@STRING{neurips = "Advances in Neural Information Processing Systems (NeurIPS)"}

@STRING{kdd = "ACM International Conference on Knowledge Discovery \& Data Mining (KDD)"}

@STRING{aaai = "Conference on Artificial Intelligence (AAAI)"}

@STRING{www = "Proceedings of the ACM on Web Conference (WWW)"}

@article{brown2020language,
  title={Language models are few-shot learners},
  author={Brown, Tom and Mann, Benjamin and Ryder, Nick and Subbiah, Melanie and Kaplan, Jared D and Dhariwal, Prafulla and Neelakantan, Arvind and Shyam, Pranav and Sastry, Girish and Askell, Amanda and others},
  journal={Advances in neural information processing systems},
  volume={33},
  pages={1877--1901},
  year={2020}
}

@inproceedings{devlin2019bert,
  title={Bert: Pre-training of deep bidirectional transformers for language understanding},
  author={Devlin, Jacob and Chang, Ming-Wei and Lee, Kenton and Toutanova, Kristina},
  booktitle={Proceedings of the 2019 conference of the North American chapter of the association for computational linguistics: human language technologies, volume 1 (long and short papers)},
  pages={4171--4186},
  year={2019}
}

@inproceedings{liu2021swin,
  title={Swin transformer: Hierarchical vision transformer using shifted windows},
  author={Liu, Ze and Lin, Yutong and Cao, Yue and Hu, Han and Wei, Yixuan and Zhang, Zheng and Lin, Stephen and Guo, Baining},
  booktitle={Proceedings of the IEEE/CVF international conference on computer vision},
  pages={10012--10022},
  year={2021}
}

@article{baevski2020wav2vec,
  title={wav2vec 2.0: A framework for self-supervised learning of speech representations},
  author={Baevski, Alexei and Zhou, Yuhao and Mohamed, Abdelrahman and Auli, Michael},
  journal={Advances in neural information processing systems},
  volume={33},
  pages={12449--12460},
  year={2020}
}

@article{li2019enhancing,
  title={Enhancing the locality and breaking the memory bottleneck of transformer on time series forecasting},
  author={Li, Shiyang and Jin, Xiaoyong and Xuan, Yao and Zhou, Xiyou and Chen, Wenhu and Wang, Yu-Xiang and Yan, Xifeng},
  journal=neurips,
  volume={32},
  year={2019}
}

@article{nie2022time,
  title={A Time Series is Worth 64Words: Long-term Forecasting with Transformers},
  author={Nie, Y},
  journal={arXiv preprint arXiv:2211.14730},
  year={2022}
}

@article{liu2023itransformer,
  title={itransformer: Inverted transformers are effective for time series forecasting},
  author={Liu, Yong and Hu, Tengge and Zhang, Haoran and Wu, Haixu and Wang, Shiyu and Ma, Lintao and Long, Mingsheng},
  journal={arXiv preprint arXiv:2310.06625},
  year={2023}
}

@inproceedings{zeng2023transformers,
  title={Are transformers effective for time series forecasting?},
  author={Zeng, Ailing and Chen, Muxi and Zhang, Lei and Xu, Qiang},
  booktitle={Proceedings of the AAAI conference on artificial intelligence},
  volume={37},
  number={9},
  pages={11121--11128},
  year={2023}
}

@article{yi2023frequency,
  title={Frequency-domain MLPs are more effective learners in time series forecasting},
  author={Yi, Kun and Zhang, Qi and Fan, Wei and Wang, Shoujin and Wang, Pengyang and He, Hui and An, Ning and Lian, Defu and Cao, Longbing and Niu, Zhendong},
  journal={Advances in Neural Information Processing Systems},
  volume={36},
  pages={76656--76679},
  year={2023}
}

@article{wu2021autoformer,
  title={Autoformer: Decomposition transformers with auto-correlation for long-term series forecasting},
  author={Wu, Haixu and Xu, Jiehui and Wang, Jianmin and Long, Mingsheng},
  journal={Advances in neural information processing systems},
  volume={34},
  pages={22419--22430},
  year={2021}
}

@inproceedings{gu2024mamba,
  title={Mamba: Linear-time sequence modeling with selective state spaces},
  author={Gu, Albert and Dao, Tri},
  booktitle={First conference on language modeling},
  year={2024}
}

@inproceedings{lea2017temporal,
  title={Temporal convolutional networks for action segmentation and detection},
  author={Lea, Colin and Flynn, Michael D and Vidal, Rene and Reiter, Austin and Hager, Gregory D},
  booktitle={proceedings of the IEEE Conference on Computer Vision and Pattern Recognition},
  pages={156--165},
  year={2017}
}

@article{azariadis1998financial,
  title={Financial intermediation and regime switching in business cycles},
  author={Azariadis, Costas and Smith, Bruce},
  journal={American economic review},
  pages={516--536},
  year={1998},
  publisher={JSTOR}
}

@article{hu2025fintsb,
  title={Fintsb: A comprehensive and practical benchmark for financial time series forecasting},
  author={Hu, Yifan and Li, Yuante and Liu, Peiyuan and Zhu, Yuxia and Li, Naiqi and Dai, Tao and Xia, Shu-tao and Cheng, Dawei and Jiang, Changjun},
  journal={arXiv preprint arXiv:2502.18834},
  year={2025}
}

@inproceedings{sun2023mastering,
  title={Mastering stock markets with efficient mixture of diversified trading experts},
  author={Sun, Shuo and Wang, Xinrun and Xue, Wanqi and Lou, Xiaoxuan and An, Bo},
  booktitle=kdd,
  pages={2109--2119},
  year={2023}
}

@inproceedings{liu2025mera,
  title={MERA: Mixture of Experts with Retrieval-Augmented Representation for Modeling Diversified Stock Patterns},
  author={Liu, YuJun and Song, Chen-Hui and Liu, Peiyuan and Li, Naiqi and Dai, Tao and Bao, Jigang and Jiang, Yong and Xia, Shu-Tao},
  booktitle=www,
  pages={1148--1152},
  year={2025}
}

@inproceedings{yoo2021accurate,
  title={Accurate multivariate stock movement prediction via data-axis transformer with multi-level contexts},
  author={Yoo, Jaemin and Soun, Yejun and Park, Yong-chan and Kang, U},
  booktitle=kdd,
  pages={2037--2045},
  year={2021}
}

@inproceedings{fan2024stockmixer,
author = {Fan, Jinyong and Shen, Yanyan},
title = {StockMixer: a simple yet strong MLP-based architecture for stock price forecasting},
year = {2024},
isbn = {978-1-57735-887-9},
publisher = {AAAI Press},
url = {https://doi.org/10.1609/aaai.v38i8.28681},
doi = {10.1609/aaai.v38i8.28681},
booktitle=aaai,
}

@inproceedings{li2024master,
  title={Master: Market-guided stock transformer for stock price forecasting},
  author={Li, Tong and Liu, Zhaoyang and Shen, Yanyan and Wang, Xue and Chen, Haokun and Huang, Sen},
  booktitle=aaai,
  volume={38},
  number={1},
  pages={162--170},
  year={2024}
}

@article{vaswani2017attention,
  title={Attention is all you need},
  author={Vaswani, Ashish and Shazeer, Noam and Parmar, Niki and Uszkoreit, Jakob and Jones, Llion and Gomez, Aidan N and Kaiser, {\L}ukasz and Polosukhin, Illia},
  journal=neurips,
  volume={30},
  year={2017}
}

@inproceedings{lin2021learning,
  title={Learning multiple stock trading patterns with temporal routing adaptor and optimal transport},
  author={Lin, Hengxu and Zhou, Dong and Liu, Weiqing and Bian, Jiang},
  booktitle=kdd,
  pages={1017--1026},
  year={2021}
}

@article{yu2024miga,
  title={MIGA: Mixture-of-Experts with Group Aggregation for Stock Market Prediction},
  author={Yu, Zhaojian and Wu, Yinghao and Wang, Genesis and Weng, Heming},
  journal={arXiv preprint arXiv:2410.02241},
  year={2024}
}

@article{feng2019temporal,
  title={Temporal relational ranking for stock prediction},
  author={Feng, Fuli and He, Xiangnan and Wang, Xiang and Luo, Cheng and Liu, Yiqun and Chua, Tat-Seng},
  journal={ACM Transactions on Information Systems (TOIS)},
  volume={37},
  number={2},
  pages={1--30},
  year={2019},
  publisher={ACM New York, NY, USA}
}

@article{hsu2021fingat,
  title={Fingat: Financial graph attention networks for recommending top-$ k $ k profitable stocks},
  author={Hsu, Yi-Ling and Tsai, Yu-Che and Li, Cheng-Te},
  journal={IEEE Transactions on Knowledge and Data Engineering (TKDE)},
  volume={35},
  number={1},
  pages={469--481},
  year={2021},
  publisher={IEEE}
}

@article{hochreiter1997long,
  title={Long short-term memory},
  author={Hochreiter, Sepp and Schmidhuber, J{\"u}rgen},
  journal={Neural computation},
  volume={9},
  number={8},
  pages={1735--1780},
  year={1997},
  publisher={MIT press}
}

@article{chung2014empirical,
  title={Empirical evaluation of gated recurrent neural networks on sequence modeling},
  author={Chung, Junyoung and Gulcehre, Caglar and Cho, KyungHyun and Bengio, Yoshua},
  journal={arXiv preprint arXiv:1412.3555},
  year={2014}
}

@article{press2021train,
  title={Train short, test long: Attention with linear biases enables input length extrapolation},
  author={Press, Ofir and Smith, Noah A and Lewis, Mike},
  journal={arXiv preprint arXiv:2108.12409},
  year={2021}
}

@article{sun2025penguin,
  title={Penguin: Enhancing transformer with periodic-nested group attention for long-term time series forecasting},
  author={Sun, Tian and Chen, Yuqi and Sun, Weiwei},
  journal={arXiv preprint arXiv:2508.13773},
  year={2025}
}

@article{raffel2020exploring,
  title={Exploring the limits of transfer learning with a unified text-to-text transformer},
  author={Raffel, Colin and Shazeer, Noam and Roberts, Adam and Lee, Katherine and Narang, Sharan and Matena, Michael and Zhou, Yanqi and Li, Wei and Liu, Peter J},
  journal={Journal of machine learning research},
  volume={21},
  number={140},
  pages={1--67},
  year={2020}
}

@article{dosovitskiy2020image,
  title={An image is worth 16x16 words: Transformers for image recognition at scale},
  author={Dosovitskiy, Alexey},
  journal={arXiv preprint arXiv:2010.11929},
  year={2020}
}

@article{tay2021pre,
  title={Are pre-trained convolutions better than pre-trained transformers?},
  author={Tay, Yi and Dehghani, Mostafa and Gupta, Jai and Bahri, Dara and Aribandi, Vamsi and Qin, Zhen and Metzler, Donald},
  journal={arXiv preprint arXiv:2105.03322},
  year={2021}
}

@article{hinton2015distilling,
  title={Distilling the knowledge in a neural network},
  author={Hinton, Geoffrey and Vinyals, Oriol and Dean, Jeff},
  journal={arXiv preprint arXiv:1503.02531},
  year={2015}
}

@inproceedings{touvron2021training,
  title={Training data-efficient image transformers \& distillation through attention},
  author={Touvron, Hugo and Cord, Matthieu and Douze, Matthijs and Massa, Francisco and Sablayrolles, Alexandre and J{\'e}gou, Herv{\'e}},
  booktitle={International conference on machine learning},
  pages={10347--10357},
  year={2021},
  organization={PMLR}
}

@inproceedings{jiao2020tinybert,
  title={Tinybert: Distilling bert for natural language understanding},
  author={Jiao, Xiaoqi and Yin, Yichun and Shang, Lifeng and Jiang, Xin and Chen, Xiao and Li, Linlin and Wang, Fang and Liu, Qun},
  booktitle={Findings of the association for computational linguistics: EMNLP 2020},
  pages={4163--4174},
  year={2020}
}

@article{bing2025optimizing,
  title={Optimizing Knowledge Distillation in Transformers: Enabling Multi-Head Attention without Alignment Barriers},
  author={Bing, Zhaodong and Li, Linze and Liang, Jiajun},
  journal={arXiv preprint arXiv:2502.07436},
  year={2025}
}

@article{liu2025efficient,
  title={Efficient multivariate time series forecasting via calibrated language models with privileged knowledge distillation},
  author={Liu, Chenxi and Miao, Hao and Xu, Qianxiong and Zhou, Shaowen and Long, Cheng and Zhao, Yan and Li, Ziyue and Zhao, Rui},
  journal={arXiv preprint arXiv:2505.02138},
  year={2025}
}

@article{liu2024large,
  title={Large language model guided knowledge distillation for time series anomaly detection},
  author={Liu, Chen and He, Shibo and Zhou, Qihang and Li, Shizhong and Meng, Wenchao},
  journal={arXiv preprint arXiv:2401.15123},
  year={2024}
}

@inproceedings{zhou2021informer,
  title={Informer: Beyond efficient transformer for long sequence time-series forecasting},
  author={Zhou, Haoyi and Zhang, Shanghang and Peng, Jieqi and Zhang, Shuai and Li, Jianxin and Xiong, Hui and Zhang, Wancai},
  booktitle={Proceedings of the AAAI conference on artificial intelligence},
  volume={35},
  number={12},
  pages={11106--11115},
  year={2021}
}

@inproceedings{xia2024ci,
  title={Ci-sthpan: Pre-trained attention network for stock selection with channel-independent spatio-temporal hypergraph},
  author={Xia, Hongjie and Ao, Huijie and Li, Long and Liu, Yu and Liu, Sen and Ye, Guangnan and Chai, Hongfeng},
  booktitle={Proceedings of the AAAI Conference on Artificial Intelligence},
  volume={38},
  number={8},
  pages={9187--9195},
  year={2024}
}

@inproceedings{tay2023scaling,
  title={Scaling laws vs model architectures: How does inductive bias influence scaling?},
  author={Tay, Yi and Dehghani, Mostafa and Abnar, Samira and Chung, Hyung and Fedus, William and Rao, Jinfeng and Narang, Sharan and Tran, Vinh and Yogatama, Dani and Metzler, Donald},
  booktitle={Findings of the Association for Computational Linguistics: EMNLP 2023},
  pages={12342--12364},
  year={2023}
}

@inproceedings{huang2022encoding,
  title={Encoding recurrence into transformers},
  author={Huang, Feiqing and Lu, Kexin and Qin, Zhen and Fang, Yanwen and Tian, Guangjian and Li, Guodong and others},
  booktitle={The Eleventh International Conference on Learning Representations},
  year={2022}
}

@inproceedings{katharopoulos2020transformers,
  title={Transformers are rnns: Fast autoregressive transformers with linear attention},
  author={Katharopoulos, Angelos and Vyas, Apoorv and Pappas, Nikolaos and Fleuret, Fran{\c{c}}ois},
  booktitle={International conference on machine learning},
  pages={5156--5165},
  year={2020},
  organization={PMLR}
}

@inproceedings{wangtimemixer,
  title={TimeMixer: Decomposable Multiscale Mixing for Time Series Forecasting},
  author={Wang, Shiyu and Wu, Haixu and Shi, Xiaoming and Hu, Tengge and Luo, Huakun and Ma, Lintao and Zhang, James Y and ZHOU, JUN},
  booktitle={The Twelfth International Conference on Learning Representations}
}

@inproceedings{xu2024rhine,
  title = {RHINE: A Regime-Switching Model with Nonlinear Representation for Discovering and Forecasting Regimes in Financial Markets},
  author = {Xu, Kunpeng and Chen, Lifei and Patenaude, Jean-Marc and Wang, Shengrui},
  booktitle = {Proceedings of the 2024 SIAM International Conference on Data Mining (SDM)},
  pages = {526--534},
  year = {2024},
  organization = {SIAM},
}

@article{andreou2002detecting,
  title={Detecting multiple breaks in financial market volatility dynamics},
  author={Andreou, Elena and Ghysels, Eric},
  journal={Journal of applied Econometrics},
  volume={17},
  number={5},
  pages={579--600},
  year={2002},
  publisher={Wiley Online Library}
}

@article{hamilton1989new,
  title={A new approach to the economic analysis of nonstationary time series and the business cycle},
  author={Hamilton, James D},
  journal={Econometrica: Journal of the econometric society},
  pages={357--384},
  year={1989},
  publisher={JSTOR}
}

@article{merton1980estimating,
  title={On estimating the expected return on the market: An exploratory investigation},
  author={Merton, Robert C},
  journal={Journal of financial economics},
  volume={8},
  number={4},
  pages={323--361},
  year={1980},
  publisher={Elsevier}
}

@article{forbes2002no,
  title={No contagion, only interdependence: measuring stock market comovements},
  author={Forbes, Kristin J and Rigobon, Roberto},
  journal={The journal of Finance},
  volume={57},
  number={5},
  pages={2223--2261},
  year={2002},
  publisher={Wiley Online Library}
}

@article{lehkonen2014timescale,
  title={Timescale-dependent stock market comovement: BRICs vs. developed markets},
  author={Lehkonen, Heikki and Heimonen, Kari},
  journal={Journal of Empirical Finance},
  volume={28},
  pages={90--103},
  year={2014},
  publisher={Elsevier}
}

@inproceedings{blondel2020fast,
  title={Fast differentiable sorting and ranking},
  author={Blondel, Mathieu and Teboul, Olivier and Berthet, Quentin and Djolonga, Josip},
  booktitle={International Conference on Machine Learning},
  pages={950--959},
  year={2020},
  organization={PMLR}
}

@inproceedings{izmailov2018averaging,
  title={Averaging weights leads to wider optima and better generalization},
  author={Izmailov, P and Wilson, AG and Podoprikhin, D and Vetrov, D and Garipov, T},
  booktitle={34th Conference on Uncertainty in Artificial Intelligence 2018, UAI 2018},
  pages={876--885},
  year={2018}
}

@article{kingma2014adam,
  title={Adam: A method for stochastic optimization},
  author={Kingma, Diederik P},
  journal={arXiv preprint arXiv:1412.6980},
  year={2014}
}
